\documentclass[]{fairmeta}
\usepackage{makecell}
\usepackage{wrapfig}
\usepackage{tabularx}
\usepackage{textcomp}
\usepackage{stfloats}
\usepackage{url}
\usepackage{verbatim}
\usepackage{titlesec}
\usepackage{tocloft}
\usepackage{adjustbox}
\usepackage{multirow}
\usepackage{pifont}
\usepackage{tikz}
\usepackage{comment}
\usepackage{amsmath,amssymb}
\usepackage{colortbl}
\usepackage{color}
\usepackage{booktabs} 
\usepackage{hyperref}
\usepackage{graphicx}
\usepackage{subcaption}
\usepackage{multirow}
\usepackage{subcaption}
\RequirePackage{xspace}
\makeatletter
\DeclareRobustCommand\onedot{\futurelet\@let@token\@onedot}
\def\@onedot{\ifx\@let@token.\else.\null\fi\xspace}
\usepackage[most]{tcolorbox}
\usepackage{xcolor}
\usepackage{array}
\usepackage{tabularx}
\usepackage{siunitx} 
\usepackage{makecell}
\usepackage[table]{xcolor}
\usepackage{caption}
\definecolor{headerpurple}{HTML}{d8d2fc}
\definecolor{rowgray}{gray}{0.95}

\def\eg{\emph{e.g}\onedot} 

\def\ie{\emph{i.e}\onedot}

\makeatother

\definecolor{adptorange}{RGB}{248, 205, 172}
\definecolor{cmpblue}{RGB}{189, 215, 238}
\definecolor{cmpblue}{RGB}{189, 215, 238}

\definecolor{our_red}{RGB}{232,157,160}
\definecolor{our_blue}{RGB}{136,206,230}
\definecolor{our_orange}{RGB}{246,200,168}
\definecolor{our_green}{RGB}{178,211,164}

\definecolor{attn_code0}{RGB}{247,215,200}
\definecolor{attn_code1}{RGB}{238,169,139}
\definecolor{mlp_code0}{RGB}{204,201,221}
\definecolor{mlp_code1}{RGB}{102,95,153}

\definecolor{token_blue}{RGB}{84, 120, 140}

\usepackage{pifont}
\usepackage{bbding}
\usepackage{fontawesome}
\usepackage{xspace}

\usepackage{float}

\newlength\savewidth

\newcolumntype{x}[1]{>{\centering\arraybackslash}p{#1pt}}
\newcolumntype{y}[1]{>{\raggedright\arraybackslash}p{#1pt}}
\newcolumntype{z}[1]{>{\raggedleft\arraybackslash}p{#1pt}}

\renewcommand{\paragraph}[1]{\vspace{1mm}\noindent\textbf{#1}}
\usepackage{colortbl}
\usepackage{xcolor}
\usepackage{wrapfig}

\renewcommand{\paragraph}[1]{\vspace{1.25mm}\noindent\textbf{#1}}

\usepackage{algorithm}
\usepackage{listings}

\definecolor{codeblue}{rgb}{0.25, 0.5, 0.5}
\definecolor{codekw}{rgb}{0.35, 0.35, 0.75}
\lstdefinestyle{Pytorch}{
    language = Python,
    backgroundcolor = \color{white},
    basicstyle = \fontsize{9pt}{8pt}\selectfont\ttfamily\bfseries,
    columns = fullflexible,
    aboveskip=1pt,
    belowskip=1pt,
    breaklines = true,
    captionpos = b,
    commentstyle = \color{codeblue},
    keywordstyle = \color{codekw},
}

\definecolor{green}{HTML}{009000}
\definecolor{red}{HTML}{ea4335}

\title{RynnVLA-001: Using Human Demonstrations to Improve Robot Manipulation}

\author[1]{Yuming Jiang}
\author[1,2]{Siteng Huang}
\author[1,2]{Shengke Xue}
\author[1,2]{Yaxi Zhao}
\author[1,2]{Jun Cen}
\author[1]{Sicong Leng}
\author[1,2]{Kehan Li}
\author[1]{Jiayan Guo}
\author[1]{Kexiang Wang}
\author[1,2]{Mingxiu Chen}
\author[1]{Fan Wang}
\author[1,2]{Deli Zhao}
\author[1,2]{Xin Li}

\affiliation[1]{DAMO Academy, Alibaba Group\\}
\affiliation[2]{Hupan Lab}

\abstract{
This paper presents \textbf{RynnVLA-001}, a vision-language-action(VLA) model built upon large-scale video generative pretraining from human demonstrations. 
We propose a novel two-stage pretraining methodology. The first stage, \textbf{Ego-Centric Video Generative Pretraining}, trains an Image-to-Video model on 12M ego-centric manipulation videos to predict future frames conditioned on an initial frame and a language instruction.
The second stage, \textbf{Human-Centric Trajectory-Aware Modeling}, extends this by jointly predicting future keypoint trajectories, 
thereby effectively bridging visual frame prediction with action prediction.
Furthermore, to enhance action representation, we propose \textbf{ActionVAE}, a variational autoencoder that compresses sequences of actions into compact latent embeddings, reducing the complexity of the VLA output space.
When finetuned on the same downstream robotics datasets,
RynnVLA-001 achieves superior performance over state-of-the-art baselines, demonstrating that the proposed pretraining strategy provides a more effective initialization for VLA models.
}

\github{\url{https://github.com/alibaba-damo-academy/RynnVLA-001}\par\vspace{0.8\baselineskip}}
\date{\today}

\begin{document}
\thispagestyle{firstheader}
\maketitle
\pagestyle{fancy}

\section{Introduction} \label{sec:introduction}
\noindent 

The past few years have witnessed rapid progress in large language models~\citep{comanici2025gemini25,anthropic2025claude4,openai2025gpt41,grattafiori2024llama3,guo2025deepseekr1,yang2025qwen3}, large multimodal models~\citep{openai2024gpt4o,bai2025qwen2,zhu2025internvl3,guo2025seed15}, vision-based recognition models~\citep{he2022mae,assran2023ijepa,oquab2023dinov2,ravi2024sam2,tschannen2025siglip2}, and generative models~\citep{Peebles:DiT,esser2024scaling,tian2024visual,baldridge2024imagen}. The success in these fields is attributed to the availability of large-scale datasets. 
For instance, large language models benefit from abundant training data readily accessible from web sources.
In contrast, progress in Vision-Language-Action (VLA) models is constrained by the scarcity of large-scale robot manipulation data. Collecting such data typically relies on human teleoperation on physical robots to record manipulation trajectories, making large-scale dataset construction both labor-intensive and costly.

There have been some early attempts to address the challenges of data scarcity. On the one hand, some methods propose to build large-scale robot manipulation datasets~\citep{O'Neill:OXE,liu2024rdt1b,khazatsky2024droid,bu2025agibot}, which collect manipulation data under diverse environments and even with different embodiments. However, the size of these datasets still remains far smaller than those used in LLMs, VLMs, and generative models~\citep{abadji2022towards,schuhmann2022laion,chen2024panda,li2024datacomp}. 
Another line of studies works on exploiting massive prior knowledge from pretrained generative models~\citep{Cheang:GR-2,Hu:VPP} or VLMs~\citep{Zitkovich:RT-2,Kim:OpenVLA,Li:CogACT,Black:pi-0,Bjorck:GR00T-N1,kim2025fine,liu2024rdt1b} to alleviate the data scarcity problem.

\begin{figure*}[t]
    \centering
    \includegraphics[width=1.0\textwidth]{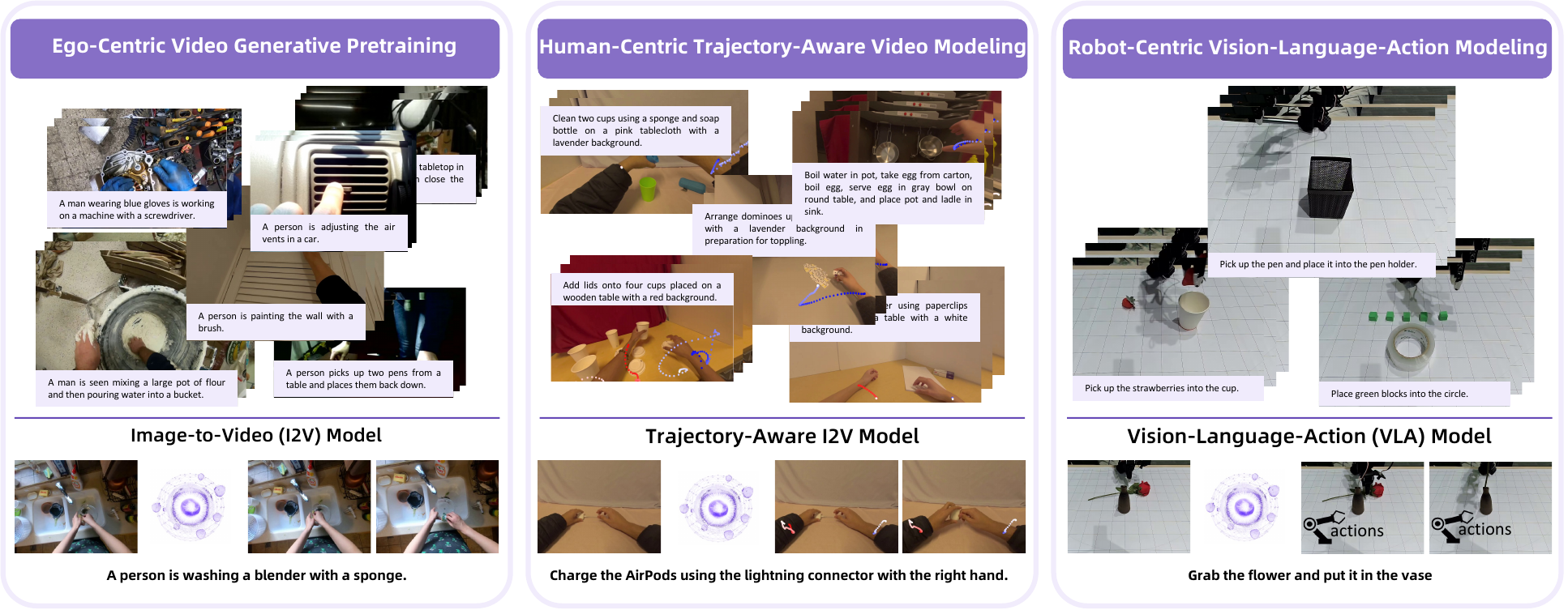}
    \vspace{0.5pt}
    \caption{
    \textbf{Training data pipeline of RynnVLA-001.} Our framework leverages three types of training data: (1) Ego-Centric Video Generative Pretraining uses millions of ego-centric human manipulation videos for future frame prediction. (2) Human-Centric Trajectory-Aware Video Modeling trains on videos with human keypoint annotations, enabling joint prediction of frames and trajectories. (3) Robot-Centric Vision-Language-Action Modeling employs robot datasets paired with language instructions to learn mappings from visual observations and language to robotic actions.
    }
    \label{fig:pipeline}
\end{figure*}

In this work, we propose RynnVLA-001, a VLA model enhanced by video generation pretraining.
The key insight of RynnVLA-001 is to implicitly transfer the manipulation skills learned from human demonstrations in ego-centric videos to robot manipulation. 
The overall training pipeline is shown in Fig.~\ref{fig:pipeline}. 
In the first stage, \textbf{Ego-Centric Video Generative Pretraining}, we train an Image-to-Video (I2V) model that takes a single image and a language instruction as inputs and predicts subsequent frames. 
To capture general manipulation dynamics, 
this stage relies on ego-centric human manipulation videos, which emphasize first-person hand operations.
We design a dedicated data curation pipeline to filter out 12M ego-centric manipulation videos from existing web sources.
Trained with ego-centric videos, the model is capable of predicting manipulations at the visual level. 
However, a gap remains between the high-level visual observations and the low-level action spaces required to control real robots.
To bridge the gap, we introduce another stage of \textbf{Human-Centric Trajectory-Aware Video Modeling}, where we further train the I2V model on ego-centric videos paired with human keypoint annotations.
In this stage, in addition to future frames, the model is also trained to predict the keypoint trajectories in future frames conditioned on current observations and language instructions.
These keypoint-based patterns share similarities with robot actions, thereby facilitating the transfer from visual dynamics to robot manipulation with low-level actions.

Following previous pretraining stages, we further adapt the model using self-collected robot datasets. 
In this phase, the model is trained to predict action chunks rather than a single-step action, conditioned on RGB observations and language instructions. To ensure the smoothness and temporal coherence of predicted actions, we propose \textbf{ActionVAE}, a variational autoencoder that encodes action chunks into compact embeddings.
Once trained, the ActionVAE is fixed and employed to extract latent representations of future actions. The model is then optimized to predict these action embeddings alongside future visual observations.
During inference, 
given an observation and a language instruction, the model outputs a single action embedding, which is subsequently decoded by ActionVAE into a sequence of executable robot actions.

Our proposed RynnVLA-001 enables a robot arm to execute complex pick-and-place and long-horizon tasks by accurately following high-level language instructions. To evaluate the effectiveness of the pretraining weights of RynnVLA-001, 
we compare our proposed RynnVLA-001 with state-of-the-art models, including GR00T N1.5~\citep{Bjorck2024GR00T15} and Pi0~\citep{Black:pi-0}, by finetuning on the same robot manipulation data.
Our proposed RynnVLA-001 consistently achieves higher success rates, demonstrating that the proposed pretraining framework provides a more effective initialization for VLA modeling.

\newpage

\section{Related Work}
\noindent 

\subsection{Vision-Language-Action Models}

The recent development of robotic agents relies on harnessing the sophisticated capabilities of pretrained Vision-Language Models (VLMs), enabling embodied systems to effectively perceive, reason, and act within complex environments.
RT-2~\citep{Zitkovich:RT-2} pioneered the concept of vision-language-action models (VLAs), which co-finetuned VLMs with both robotic trajectory data and large-scale Internet vision-language tasks.
A key strategy in this line of work was the discretization of continuous actions into categorical bins, making them compatible with standard LLM backbones.
This approach was further advanced by models like OpenVLA~\citep{Kim:OpenVLA}, which improved generalization through extensive pretraining, and FAST~\citep{Pertsch:FAST}, which explored more efficient tokenization schemes.
However, action discretization often leads to precision loss in action representations.
VQ-VLA~\citep{wang2025vq} proposed to train a VQVAE to discretize the action chunks, which improved the accuracy of action representation.
Another line of approaches proposed to integrate policy heads specifically designed for continuous action generation.
LCB~\citep{Shentu:LCB} pioneered such a dual-system architecture by leveraging a pretrained VLM~\citep{Liu:LLaVA} for high-level reasoning and scene understanding, while employing a separate and specialized policy head to generate the continuous and low-level actions~\citep{Ke:3D-Diffuser-Actor}.
Numerous subsequent studies have built upon this dual-system paradigm, exploring various refinements such as optimizing interaction frequencies, employing different policy head models like diffusion transformers~\citep{Peebles:DiT}, and incorporating diverse training strategies across multiple embodiments~\citep{Zhang:HiRT,Li:CogACT,Wen:DexVLA,Lin:OneTwoVLA,Cui:Open-Helix,Shukor:SmolVLA,Cheang:GR-3}.
Notably, recent efforts have matured this concept into more general frameworks. For instance, 
Pi0~\citep{Black:pi-0} integrated PaliGemma~\citep{Beyer:PaliGemma} with conditional flow matching~\citep{Lipman:flow-matching} for continuous control.
Similarly, the GR00T project~\citep{Bjorck:GR00T-N1, Bjorck2024GR00T15} has delivered an end-to-end trained and open-source dual-system on complex hardware like humanoid robots.
These frameworks represent a significant step towards advanced VLM-based VLA models.

\subsection{Future Prediction for Robot Learning}

Existing VLM-based policies are often limited by inadequate modeling of visual dynamics, as they typically rely on one or two static images from the current observation. To address this, a significant stream of research has focused on incorporating future prediction to explicitly model physical dynamics and thereby improve policy learning. These efforts can be broadly categorized into three paradigms.
The first paradigm conditions the policy on an explicitly generated future state. Early works like SuSIE~\citep{Black:SuSIE} and UniPi~\citep{Du:UniPi} leveraged image editing or generation models to produce a single future keyframe, which then served as a visual goal for the action policy. This concept was later advanced by methods such as GEVRM~\citep{Zhang:GEVRM}, which utilized powerful video foundation models like Open-Sora~\citep{Zheng:Open-Sora} to generate more expressive future goal sequences. DREAMGEN~\citep{Jang2025DreamGenUG} employed video world models to generate large-scale synthetic robot data for offline policy training.
A more integrated paradigm unifies future prediction and action generation within a single architecture. Some approaches~\citep{Wu:GR-1,Tian:Seer,Zhao:CoT-VLA} enforced a sequential dependency to generate a future subgoal image before predicting the action sequence. 
PAD~\citep{Guo:PAD} employed a diffusion model to jointly forecast future images and actions, while WorldVLA~\citep{Cen:WorldVLA} developed an autoregressive framework where multi-step future state prediction and action generation mutually inform and benefit each other.
The third paradigm leverages future video prediction primarily as a powerful pretraining objective for representation learning, rather than as a direct component during inference. For instance, GR-2~\citep{Cheang:GR-2} pretrained a model on vast internet video datasets before finetuning for action prediction, while VPP~\citep{Hu:VPP} successfully leveraged representations from pre-existing video foundation models for robotic control.
While these methods validate the potential of video pretraining, our work advances this paradigm by introducing a multi-stage curriculum that progressively connects visual generation with robot control. 
We first pretrain a video generation model on large-scale ego-centric videos. 
Next, we introduce an intermediate stage that incorporates human trajectory prediction to bridge the gap between purely visual prediction and action generation.
Finally, this trajectory-aware model is finetuned on robot-specific data.

\section{Methodology}

\begin{figure*}[t]
    \centering
    \includegraphics[width=1.0\textwidth]{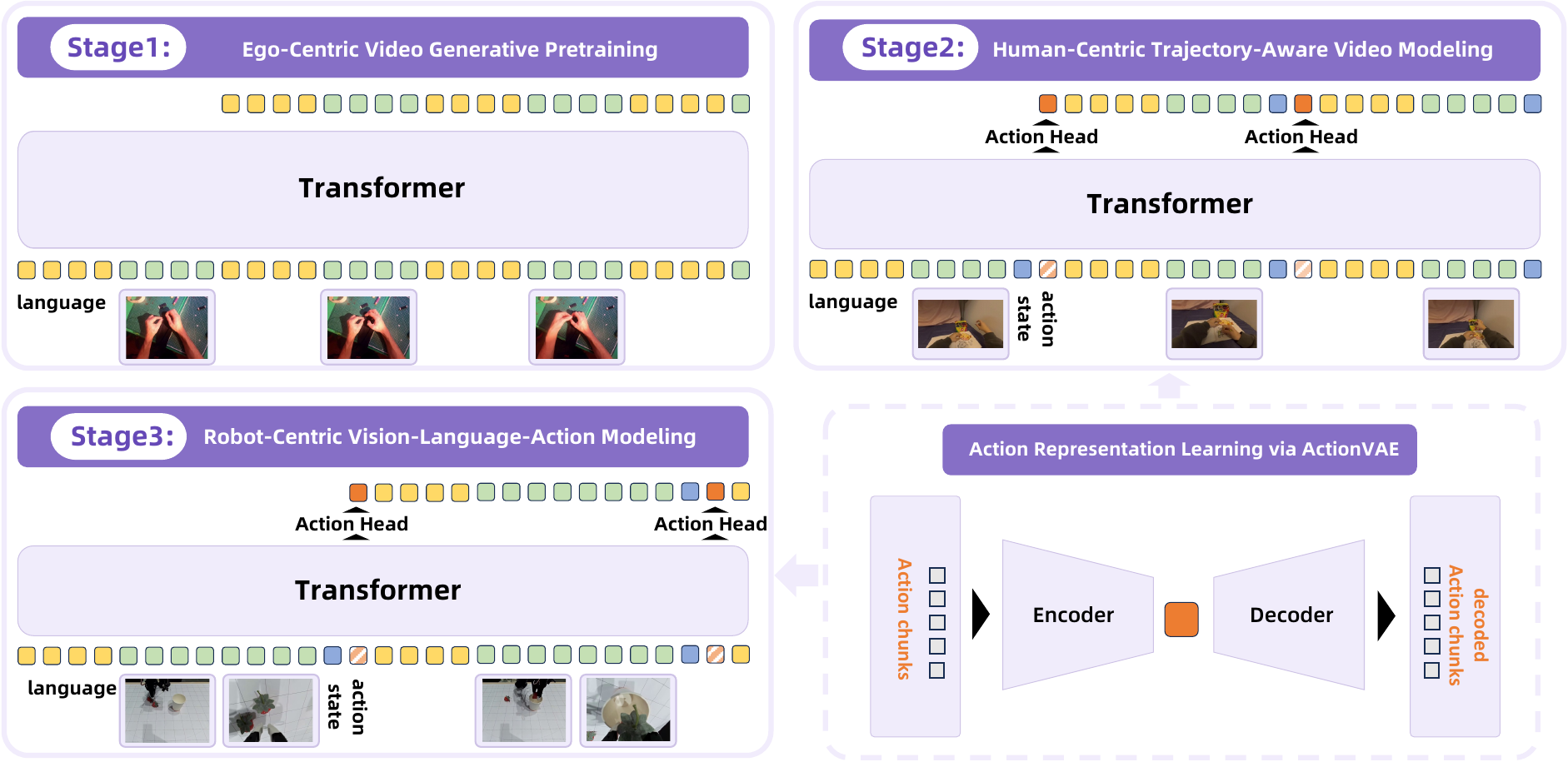}
    \vspace{0.5pt}
    \caption{
    \textbf{Model architecture and training stages of RynnVLA-001.} The training consists of three stages: (1) Ego-Centric Video Generative Pretraining trains a transformer-based Image-to-Video (I2V) model for future frame prediction. (2) Human-Centric Trajectory-Aware Video Modeling extends the I2V model with action (trajectory) prediction heads, incorporating both visual and state embeddings (blue blocks). (3) Robot-Centric Vision-Language-Action Modeling transfers pretrained weights to robot data, where the model generates action embeddings decoded by ActionVAE into executable actions.
    }
    \label{fig:architecure}
\end{figure*}

\noindent
In this work, we propose RynnVLA-001, a Vision-Language-Action (VLA) model built upon large-scale video generation pretraining. We first train an ego-centric Image-to-Video (I2V) model, then finetune it into a trajectory-aware ego-centric video generation model. Finally, we adapt the model into a VLA model by inheriting the pretrained weights.
The model is progressively trained through three stages, as illustrated in Fig.~\ref{fig:architecure}:
\textbf{1) Ego-Centric Video Generative Pretraining}: An ego-centric Image-to-Video (I2V) model is trained on ego-centric human manipulation videos. This stage enables the model to predict future frames. 
\textbf{2) Human-Centric Trajectory-Aware Video Modeling}: The pretrained I2V model is further finetuned to jointly predict future frames and human keypoint trajectories. This stage bridges the gap between purely visual frame prediction and action-oriented modeling.
\textbf{3) Robot-Centric Vision-Language-Action Modeling}: The VLA model inherits the weights from the previous stages and is trained on robot data using language instructions and current observations (including two-view observations and joint states) as inputs. It is optimized to predict both future frames and action embeddings, with the latter decoded by the pretrained ActionVAE into executable robot actions.

\subsection{Ego-Centric Video Generative Pretraining}
A key challenge in scaling VLA models is the scarcity of large-scale paired training data. To address this, we transfer priors from human demonstrations through large-scale video pretraining, beginning with Ego-Centric Video Generative Pretraining.

The objective of this stage is to train an I2V model that closely mimics the inference process of a VLA model. In a typical VLA setting, actions are predicted conditioned on current observations (\eg, visual inputs and robot states) and a language instruction. Accordingly, our I2V model is trained to predict future video frames based on an initial visual observation and a corresponding language description. This pretraining task forces the model to learn the physical dynamics of object manipulation from an ego-centric perspective.

For the network architecture, we adopt an autoregressive (AR) Transformer. Due to the limited availability of AR-based video generation frameworks, we extend a powerful AR image generation model, Chameleon~\citep{Chameleonpaper, lumina-mgpt}, to perform the I2V task. To ensure the learned priors are directly relevant for robotic manipulation, 
we curate $12$M ego-centric human manipulation videos for training. These videos contain first-view human operations and focus on hand manipulations that are analogous to robot gripper movements and operations (Details of the data curation pipeline will be provided in Sec.~\ref{sec:data_curation}.). Besides, we also filter $244$K robotic manipulation videos from open-source datasets~\citep{brohan2022rt, walke2023bridgedata, O'Neill:OXE, khazatsky2024droid}.
As shown in Fig.~\ref{fig:architecure}, instead of providing the instruction only once, language tokens are interleaved with visual tokens to better align with VLA inference, where each action prediction is conditioned on both modalities. The input sequence is structured as
\begin{equation*}
    [\texttt{language tokens}, \texttt{visual tokens}_{t}, \texttt{language tokens}, \texttt{visual tokens}_{t+1}, ...].
\end{equation*}
The training is supervised by the cross-entropy loss over discrete visual tokens and language tokens.

\subsection{Human-Centric Trajectory-Aware Video Modeling}
While the I2V model from Stage 1 learns to predict dynamics at the visual level, it lacks an explicit understanding of actions, which is essential for the final VLA modeling. To bridge the gap between purely visual prediction and action generation, Stage 2 refines the pretrained model into a trajectory-aware framework. 

The key idea of this stage is to finetune the pretrained model with a multi-task objective: to concurrently predict future visual frames and the corresponding human keypoint trajectories. Human trajectories can be regarded as another form of actions. By learning to associate visual changes with their underlying motion trajectories, the model develops a more holistic understanding of manipulation dynamics.

To this end, we adopt the EgoDex dataset~\citep{egodex_paper}, which provides trajectories of all upper-body joints captured via Apple Vision Pro devices. Among these, we selectively use only the wrist keypoints, as they approximate end-effector positions. Crucially, rather than predicting raw coordinates, the model is trained to predict compact and continuous embeddings of trajectory chunks generated by a pretrained ActionVAE (detailed in Sec.~\ref{sec:actionvae}). This design yields a dense representation of motion while reducing prediction complexity.

To provide the model with proprioceptive information, we introduce state embeddings (blue blocks in Fig.~\ref{fig:architecure}). These embeddings represent the current keypoint positions of the human wrists and are fed into the model at each timestep. 
These are projected into the transformer’s input dimension via a linear layer and interleaved with other tokens.
The input sequence is now structured as: 
\begin{equation*}
[\texttt{language}, \texttt{visual tokens}_t, \texttt{state embedding}_t, \texttt{<ACTION\_PLACEHOLDER>}, ...], 
\end{equation*}
where \texttt{<ACTION\_PLACEHOLDER>} is the signal to generate continuous action embeddings. 
This sequence explicitly provides the model with three crucial pieces of information: the high-level goal (language), the current visual scene (visual tokens), and the current physical configuration of the wrists (state embeddings). 

The architecture from Stage 1 is extended to handle the new and continuous prediction target. 
A lightweight action head (a single linear layer) is introduced. The transformer's main output remains discrete visual tokens, while the action head maps the last hidden state to the continuous latent space of action embeddings. 
The training of the action head is supervised by L1 loss, which is computed exclusively for the outputs at token positions of \texttt{<ACTION\_PLACEHOLDER>}.
The prediction of visual tokens remains the same as Stage 1.

\subsection{ActionVAE: Action Representaton via VAE}\label{sec:actionvae}
In VLA models, predicting action chunks (\ie, short sequences of actions) is more effective than predicting single and step-by-step actions~\citep{zhao2023learning, kim2025fine}. This design choice is motivated by two key factors: 1) Avoiding repetitive predictions: Single-step action execution often results in negligible visual changes, which can cause the model to repeatedly output the same action and become stuck. Predicting a chunk encourages more substantial visual changes. 2) Efficiency: Generating multiple actions in a single forward pass reduces computational overhead and inference latency.

To facilitate this chunk-level prediction while ensuring the generated actions are smooth and coherent, we introduce an Action Variational Autoencoder (ActionVAE) in Stage 2\&3. As illustrated in Fig.~\ref{fig:architecure}, the ActionVAE consists of an encoder that compresses an ``action chunk'' into a compact and continuous latent embedding, and a decoder that reconstructs the original action sequence from this embedding.

Since our training pipeline involves both human demonstrations and robot executions, and their kinematic spaces differ, we train two domain-specific ActionVAEs:
one for compressing human trajectories (used in Stage 2) and another for compressing robot actions (used in Stage 3). This ensures that each domain has a tailored and accurate action representation.
Importantly, the ActionVAE is embodiment-specific. As it encodes chunk-level actions that often correspond to atomic motion primitives, a well-trained model can be directly used to extract action embeddings from new data on the same embodiment without retraining\footnote{
Our pretrained ActionVAE can extract action embeddings from unseen data. Our demo video demonstrates this: although the ActionVAE was never trained on tasks such as \texttt{pick up the small cube and put it on the big cube}, \texttt{pick up the holder and place it straight then pick up the pen and place it in the holder}, \texttt{Tower of Hanoi with two disks}, and \texttt{put the strawberry and solid glue on the palm}, the embeddings it extracts can be used to train the downstream VLA model.}.

\subsection{Robot-Centric Vision-Language Action Modeling}
In the final stage, we adapt the pretrained trajectory-aware model into a VLA model for robot control. This is achieved by integrating the robot-specific ActionVAE representations and finetuning the model on robot-centric data. The model's primary objective is to predict the embedding of the next robot action chunk, which is then decoded by the ActionVAE into an executable action sequence.

The architecture largely inherits the framework from Stage 2 but is now adapted for the robot domain. A key modification lies in the action head. 
Since human hand trajectories differ substantially from robot arm kinematics, the action head pretrained in Stage 2 is discarded. Instead, a new lightweight action head (a single linear layer) is initialized to predict robot action embeddings.

The input sequence for the VLA model is structured to mirror the real-world robotic deployment scenario, using the same placeholder-based design as Stage 2: 
\begin{equation*}
[\texttt{language}, \texttt{visual tokens}_t, \texttt{state embedding}_t, \texttt{<ACTION\_PLACEHOLDER>}, ...], 
\end{equation*}
where the visual tokens now comprise two camera views (a front view and a wrist view), as opposed to the single-view input in Stage 2. This sequence explicitly provides four components: (1) the high-level goal (language), (2) robot-centric visual observations (front and wrist views), (3) the current robot state, and (4) a signal to generate the next action (\texttt{<ACTION\_PLACEHOLDER>}).

During training, the model is optimized with two concurrent objectives: 1) Robot Action Prediction: The hidden state corresponding to the output of the \texttt{<ACTION\_PLACEHOLDER>} token is fed into the newly initialized action head. This head regresses the hidden state to a continuous embedding representing the next robot action chunk. The training of this head is supervised by an L1 loss between the predicted embedding and the ground-truth embedding from the robot-specific ActionVAE. 2) Future Visual Prediction: The model continues the autoregressive prediction of visual tokens for the next frame, supervised by cross-entropy loss. This auxiliary task regularizes training and preserves the model’s understanding of world dynamics.

\subsection{Inference}
During inference, the VLA model operates within a closed-loop control cycle to perform tasks. At each step of the cycle, the model receives the language instruction, the current RGB observation from the robot's cameras and current robot states as inputs.

Crucially, to optimize for efficiency, we make a modification at inference time. The model only predict the action embedding, and discards the generation of future vision tokens. While predicting future frames serves as a valuable auxiliary task for regularization during training, it is computationally expensive and unnecessary for control. Discarding this process significantly increases inference speed, making the model more practical for real-time applications.

The predicted action embedding is then immediately passed to the decoder of the ActionVAE. The decoder reconstructs a coherent sequence (or chunk) of low-level robot actions from this single embedding. The robot executes this entire action chunk. Upon completion, a new observation is captured and fed back into the model along with the language instruction, initiating the next cycle. This process repeats until the task is successfully completed.

\section{Ego-Centric Video Data Curation Pipeline}
\label{sec:data_curation}

To pretrain our model on relevant human manipulation demonstrations, we construct a large-scale dataset of ego-centric manipulation videos from web sources~\citep{wang2024egovid, grauman2022ego4d, miech19howto100m, Damen2022RESCALING, Damen2018EPICKITCHENS, Damen2021PAMI, goyal2017something, mahdisoltani2018effectiveness}. Since raw video data is noisy and highly diverse, we design a multi-stage data curation pipeline to filter and annotate videos suited for our pretraining stage. The pipeline consists of the following steps: 

\noindent\textbf{Keypoint Detection.}
For each frame in the video, we apply a pose estimation model~\citep{yang2023effective} to extract human keypoints, including facial landmarks, torso joints, and hand-related keypoints (\eg, wrists, elbows, fingers). 

\noindent\textbf{Ego-centric Filtering.}
We apply two key filtering criteria to retain only high-quality ego-centric manipulation videos: 1) No facial keypoints: Videos containing facial landmarks are discarded. The appearance of a human face strongly suggests a third-person perspective, which is not suitable for ego-centric modeling. 2) Presence of hand keypoints: We retain only frames where keypoints corresponding to the wrists and hands are visible. The presence of hands near the camera is a strong indicator of ego-centric human manipulation, which is essential for modeling transferable robotic behavior.

\noindent\textbf{Text Description Annotation.}
For each curated video, we use the Qwen2-VL-7B~\citep{Qwen2VL} to generate concise textual descriptions. These descriptions are intentionally kept short to mimic natural language instructions typically used in robot learning (\eg, ``put the bottle in the box'', ``open the drawer''). The resulting text-video pairs serve as effective supervision signals for vision-action alignment.
\begin{figure*}[t]
    \centering
    \includegraphics[width=1.0\textwidth]{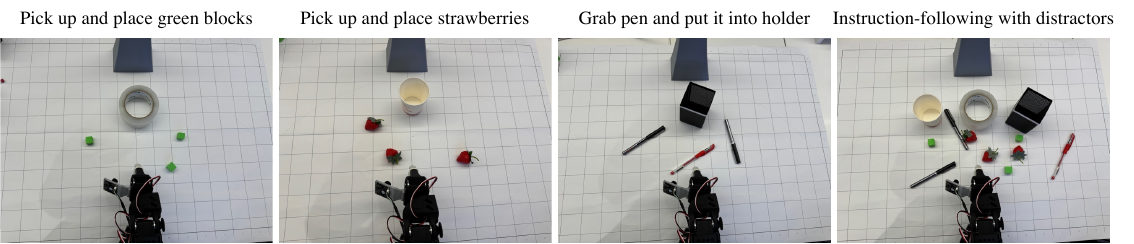}
    \vspace{0.3pt}
    \caption{\textbf{Illustration of Evaluation Tasks.} We evaluate the performance of VLA models on three tasks: (1) pick up and place green blocks, (2) pick up and place strawberries, and (3) grab pen and put it into holder. Each task is evaluated under three settings: single-target manipulation, multi-target manipulation (first three images), and instruction-following with distractors (rightmost image).}
    \label{fig:task_illustration}
\end{figure*}

\section{Experiments}
\noindent

\subsection{Experimental Setup}

\noindent\textbf{Dataset.} To train and evaluate our proposed RynnVLA-001 model, we collect a new real-world manipulation dataset using a LeRobot SO100 robotic arm~\citep{cadene2024lerobot}. The dataset comprises expert demonstrations collected through human teleoperation. To ensure our dataset covers basic manipulation skills, we design and collect data for three representative tasks as shown in Fig.~\ref{fig:task_illustration}: 1) \texttt{Pick up and place green blocks}: This task focuses on fundamental object recognition and grasping abilities. We collect 248 demonstrations. 2) \texttt{Pick up and place strawberries}: This task requires precise localization and grasping point estimation, focusing on the model's fine-grained perception capabilities. We collect 249 demonstrations. 3) \texttt{Grab pen and put it into holder}: This task demands advanced 3D spatial reasoning, specifically the ability to infer object orientation and height for a precise insertion action. We collect 301 demonstrations.
To enhance the richness and complexity of the data, the scenes of manipulation are set to vary from containing only target objects to more complex arrangements that include other irrelevant, distractor objects.
During teleoperation, the human operator's goal is to move all target objects to their destination.
Furthermore, the data is collected using three different SO100 arms in various environments with different lighting conditions.

\noindent\textbf{Baselines.}
We compare our model with two strong open-source baselines, namely GR00T N1.5~\citep{Bjorck2024GR00T15} and Pi0~\citep{Black:pi-0}. We initialize these models from the corresponding pretrained weights and then finetune the model with the same SO100 data as our model. We use the official code of GR00T N1.5 and Pi0 and strictly follow the instructions to finetune the model.

\begin{table}[t]
\centering
\caption{\textbf{Performance comparison on three manipulation tasks.} We report task-specific success rates, average success rate over three tasks, and SR@1. Each number represents the average SR across the three evaluation settings in Tab.~\ref{tab:diff_eval}.}
\small
\renewcommand{\arraystretch}{1.3}

\begin{tabular}{
  @{} l 
  S[table-format=2.1, detect-weight]
  S[table-format=2.1, detect-weight]
  S[table-format=2.1, detect-weight]
  S[table-format=2.1, detect-weight]
  S[table-format=2.1, detect-weight] @{}
}
\toprule
\rowcolor{headerpurple!60}
\multicolumn{1}{@{}c}{\multirow{2}{*}{\textbf{Method}}} & \multicolumn{3}{c}{\textbf{Task-Specific Success Rate (\%)}} & {\multirow{2}{*}{\textbf{Average (\%)}}} & {\multirow{2}{*}{\textbf{SR@1 (\%)}}} \\
[4pt]
\rowcolor{headerpurple!60}
& {\thead{Pick up and place \\ green blocks}} & {\thead{Pick up and place \\ strawberries}} & {\thead{Grab pen and \\ put it into holder}} & & \\
\midrule
GR00T N1.5 & 65.0 & 53.3 & 48.3 & 55.6 & 37.2 \\
Pi0 & 75.6 & 71.1 & 64.4 & 70.4 & 56.3 \\
\bfseries RynnVLA-001 (Ours) & \bfseries 90.0 & \bfseries 91.7 & \bfseries 90.0 & \bfseries 90.6 & \bfseries 56.7 \\
\bottomrule
\end{tabular}
\label{tab:sota_compare_task}
\end{table}

\begin{table}[t]
\centering
\caption{\textbf{Performance comparison on three different evaluation settings.} Single-target manipulation refers to the setting only a single target object present on the desktop. Multi-target manipulation means multiple target objects present on the desktop. Instruction-following with distractors refers to the test cases where target objects and distractor objects are present on the desktop at the same time.}
\label{tab:diff_eval}
\small
\renewcommand{\arraystretch}{1.3}

\begin{tabular}{
  @{} l c c c @{}
}
\toprule
\rowcolor{headerpurple!60}
{\textbf{Method}} & 
{\textbf{Single-Target Manipulation}} &  
{\textbf{Multi-Target Manipulation}} &
{\textbf{\thead{Instruction-following with Distractors}}} \\
\midrule
GR00T N1.5 & 63.3 & 46.7 & 56.7 \\
Pi0 & 80.0 & 71.1 & 60.0 \\
\bfseries RynnVLA-001 (Ours) & \bfseries 93.3 & \bfseries 86.7 & \bfseries 91.7  \\
\bottomrule
\end{tabular}
\label{tab2:comparison_settings}
\end{table}

\noindent\textbf{Evaluation.}
We consider three different scenarios in our evaluation protocol: 1) \texttt{Single-target Manipulation}, where only a single target object is on the desktop,  2) \texttt{Multi-target Manipulation}, where multiple target objects are on the desktop, 3) \texttt{Instruction-following with Distractors}, where both target objects and distractor objects appear on the desktop.
For all scenarios, a trial is considered a success if the model correctly places at least one target object in its designated location within a predefined time limit.
A trial is marked as a failure under any of the following conditions: 1) The time limit is exceeded. 2) The model makes more than five consecutive failed attempts to grasp a target object. 3) Specifically, for the Instruction-Following with Distractors scenario, the model attempts to manipulate any distractor objects.
We also report the Success Rate@1 (SR@1) metric, defined as the percentage of tasks successfully completed within a single trial.
To evaluate generalization, each task is evaluated on multiple robotic arms, each operating in a unique physical environment.

\subsection{Comparison with SoTA methods}

Table~\ref{tab:sota_compare_task} presents a detailed comparison of task-specific and average success rates. Our model, RynnVLA-001, demonstrates substantially higher overall performance, outperforming both GR00T N1.5 and Pi0 across all three tasks. As for the SR@1 metric, the performance of our proposed RynnVLA-001 is comparable to Pi0. 
The relatively low SR@1 values for all three models suggest that there is still a large zoom for improving object localization accuracy to achieve reliable single-trial success.

In Table~\ref{tab2:comparison_settings}, we report success rates in three different evaluation settings. The task becomes more challenging when more objects appear in the scene. For GR00T N1.5, the success rates of multi-target manipulation and instruction-following with distractor objects become lower than those of single-target manipulation. For Pi0, when distractor objects appear on the desk, the success rates drop significantly, indicating its limited instruction-following capability.
In contrast, the performance of our proposed RynnVLA-001 remains stable.

\begin{table}[t]
\centering
\caption{\textbf{Effectiveness of Pretrained Weight on three evaluation settings.} We train four variants of RynnVLA-001 with four different initialized weights. We report the success rates on three tasks, average success rate and SR@1.}
\small
\renewcommand{\arraystretch}{1.3}

\begin{tabular}{
  @{} l
  S[table-format=2.1, detect-weight]
  S[table-format=2.1, detect-weight]
  S[table-format=2.1, detect-weight]
  S[table-format=2.1, detect-weight]
  S[table-format=2.1, detect-weight] @{}
}
\toprule
\rowcolor{headerpurple!60}
\multicolumn{1}{@{}c}{\multirow{2}{*}{\textbf{Method}}} & \multicolumn{3}{c}{\textbf{Task-Specific Success Rate (\%)}} & {\multirow{2}{*}{\textbf{Average (\%)}}} & {\multirow{2}{*}{\textbf{SR@1 (\%)}}} \\
[4pt]

\rowcolor{headerpurple!60}
& {\thead{Pick up and place \\ green blocks}} & {\thead{Pick up and place \\ strawberries}} & {\thead{Grab pen and \\ put it into holder}} & & \\
\midrule
RynnVLA-001-Scratch\footnotemark & 0 & 6.7 & 6.7 & 4.4 & 0 \\
RynnVLA-001-Chameleon & 56.6 & 50.0 & 43.3 & 50.0 & 22.8 \\
RynnVLA-001-Video & 81.7 & 86.7 & 85.0 & 84.4 & 49.4 \\
\bfseries RynnVLA-001 (Full) & \bfseries 90.0 & \bfseries 91.7 & \bfseries 90.0 & \bfseries 90.6 & \bfseries 56.7 \\
\bottomrule
\end{tabular}
\label{tab3:pretrained_success}
\end{table}

\footnotetext{Given the low performance of the RynnVLA-001-Scratch model, its evaluation is limited to 5 trials per task and setting, and conducted on a single robot arm. In contrast, all other models are evaluated with 60 trials per task. These trials are distributed evenly across two robotic arms, with each arm conducting 10 trials for each of the three scenarios (totaling 30 trials per arm).}

\begin{table}[t]
\centering
\caption{\textbf{Effectiveness of Pretrained Weight.} We train four variants of RynnVLA-001 with four different initialized weights. We report the success rates in three different evaluation settings.}
\small
\renewcommand{\arraystretch}{1.3}

\begin{tabular}{
  @{} l c c c @{}
}
\toprule
\rowcolor{headerpurple!60}
{\textbf{Method}} & 
{\textbf{Single-Target Manipulation}} & 
{\textbf{Multi-Target Manipulation}} &
{\textbf{\thead{Instruction-Following with Distractors}}} \\
\midrule
RynnVLA-001-Scratch & 0 & 6.7 & 6.7 \\
RynnVLA-001-Chameleon & 55.0 & 53.3 & 41.7 \\
RynnVLA-001-Video & 93.3 & 78.3 & 81.7 \\
\bfseries RynnVLA-001 (Full) & \bfseries 93.3 & \bfseries 86.7 & \bfseries 91.7  \\
\bottomrule
\end{tabular}
\label{tab4:pretrained_different_settings}
\end{table}

\subsection{Effectiveness of Pretraining}

In RynnVLA-001, we propose two pretraining stages: 1) ego-centric video generative pretraining and 2) human-centric trajectory-aware video modeling. To investigate the effectiveness of our proposed two-stage pretraining pipeline, we conduct a comprehensive ablation study, with results presented in Table~\ref{tab3:pretrained_success} and Table~\ref{tab4:pretrained_different_settings}.

First, we evaluate the impact of Stage 1: Ego-centric Video Generative Pretraining. We compare three initialization strategies for the final VLA model:
1) RynnVLA-001-Scratch: A baseline initialized from random weights, skipping all pretraining.
2) RynnVLA-001-Chameleon: A stronger baseline initialized directly from the pretrained weights of the Chameleon Text-to-Image (T2I) model~\citep{Chameleonpaper}, bypassing our video pretraining stage.
3) RynnVLA-001-Video: Our model after the training of Stage 1, which starts from Chameleon weights but is further pretrained on ego-centric videos.
The results clearly demonstrate the importance of video-centric pretraining. The RynnVLA-001-Scratch model is incapable of correlating language instructions with meaningful actions, resulting in an extremely low success rate. 
Benefiting from the pretrained T2I checkpoint, the RynnVLA-001-Chameleon model achieves reasonable results on simple grasping.
However, it exhibits a limited localization capability, capping its performance at a success rate of 50.0\%. In contrast, RynnVLA-001-Video achieves a significant performance improvement, indicating that priors learned from ego-centric videos are effective for VLA adaptation.

Next, we build upon this video-pretrained model to evaluate the contribution of Stage 2: Human-centric Trajectory-Aware Video Modeling. By incorporating this second pretraining stage where the model learns to predict human trajectories, our full model, RynnVLA-001, achieves the best performance among all variants. This final improvement shows the benefit of explicitly bridging the gap between visual prediction and action generation by pretraining the model to predict human trajectories.

\begin{table}[ht]
\centering
\caption{\textbf{Ablation Study of VLA Components on the Calvin Benchmark.} All models are trained with reduced epochs for efficiency; scores are for relative comparison.}
\small
\renewcommand{\arraystretch}{1.3}

\begin{tabular}{
  @{} l 
  l    
  S[table-format=2.1]
  S[table-format=2.1]
  S[table-format=2.1]
  S[table-format=2.1]
  S[table-format=2.1]
  S[table-format=1.3] @{} 
}
\toprule
\rowcolor{headerpurple!60}
\multicolumn{1}{@{}l}{\multirow{2}{*}{\textbf{Method}}} &
\multicolumn{1}{l}{\multirow{2}{*}{\textbf{Task}}} &
\multicolumn{5}{c}{\textbf{Task Success Rate (\%)}} &
\multicolumn{1}{c@{}}{\multirow{2}{*}{\textbf{Avg. Len.}}} \\
[4pt]

\rowcolor{headerpurple!60}
\multicolumn{1}{@{}l}{} &
\multicolumn{1}{l}{} &  
1 & 2 & 3 & 4 & 5 &
\multicolumn{1}{c@{}}{} \\ 
\midrule

256 x 256 & Task ABC -> D & 92.7 & 83.7 & 73.5 & 62.1 & 53.2 & 3.652 \\
\hline 
Raw Actions Prediction & Task ABC -> D & 93.8 & 86.5 & 80.4 & 74.2 & 67.0 & 4.019 \\
\hline
Deeper Action Head & Task ABC -> D & 90.2 & 77.9 & 65.3 & 54.6 & 44.3 & 3.323 \\
\hline
\bfseries Full Model &  Task ABC -> D & \bfseries{95.4} & \bfseries{88.2} & \bfseries{82.2} & \bfseries{78.2} & \bfseries{72.1} & \bfseries{4.161} \\
\bottomrule
\end{tabular}
\label{tab:ablation}
\end{table}

\subsection{Ablation Study on Model Designs}

To systematically evaluate the impact of the developed key components, we conduct a series of ablation studies on the Calvin Benchmark~\citep{mees2022calvin}. For experimental efficiency, our full model and the ablated variants are trained from the pretraining weights of RynnVLA-001-Video but for a reduced number of epochs. \footnote{We modify the evaluation for the ``place in slider'' task because the original prompt, ``store the grasped block in the sliding cabinet'', results in extremely low performance. To better assess action prediction capabilities, we revise it to ``place the grasped object in the sliding cabinet''. }
Consequently, the results presented in this section are intended for comparative analysis to demonstrate the relative importance of each component.

\noindent\textbf{Image Resolution.}
In this study, we investigate the impact of image resolution on RynnVLA-001-Video.
As shown in Tab.~\ref{tab:ablation}, a substantial performance drop is observed when the resolution decreases from our proposed $384 \times 384$ to $256 \times 256$. This degradation is attributed to the resolution mismatch with the VQGAN component, which is pretrained exclusively on $512 \times 512$ images. At a lower resolution of $256 \times 256$, the VQGAN's reconstruction quality degrades, the VQGAN fails to generate high-fidelity reconstructions, resulting in imprecise visual tokens that cannot faithfully represent the source content. Consequently, a VLA model trained on these imprecise tokens exhibits reduced performance. Furthermore, our choice of $384 \times 384$ strikes a balance: 1) it maintains high reconstruction fidelity by using the resolution closer to the VQGAN's native resolution; 2) it offers a significant reduction in computational overhead compared to the $512 \times 512$ resolution, making it a more practical choice for deployment.

\noindent\textbf{Action Representations.} 
In this work, we propose to use a Variational Autoencoder (VAE) to compress action chunks into compact latent embeddings. This approach contrasts with prior methods~\citep{Kim:OpenVLA, Black:pi-0, Bjorck:GR00T-N1} that directly predict raw action sequences. To evaluate the effectiveness of the component, we conduct an ablation study on the Calvin ABC->D benchmark, comparing the performance of predicting VAE embeddings against predicting raw actions.
As shown in Tab.~\ref{tab:ablation}, predicting actions in the VAE's latent space outperforms the direct prediction of raw actions. The performance gains stem from two key advantages provided by the VAE: 1) it provides an efficient, compressed representation of complex action sequences, and 2) the inherent structure of its latent space improves temporal consistency, yielding smoother predicted actions.

\noindent\textbf{Size of Action Head.}
Our action prediction module utilizes a simple action head, \ie, a single linear layer that projects the transformer's final hidden state into the action embedding space. To assess the impact of head complexity, we perform an ablation study comparing this design with a deeper five-layer MLP head on the Calvin Task ABC->D benchmark.
As shown in Tab.~\ref{tab:ablation}, increasing the depth of action head is surprisingly detrimental to performance, causing the evaluation score to decrease substantially from 4.019 to 3.323. This result indicates that the transformer's output representation is already highly effective for the task. A direct linear mapping is sufficient for decoding, while the additional complexity of a deeper head appears to introduce noise or overfitting, ultimately impairing performance. This underscores the value of architectural simplicity in the decoding stage of our model.

\begin{figure*}[htbp]
    \centering
    \includegraphics[width=1.0\textwidth]{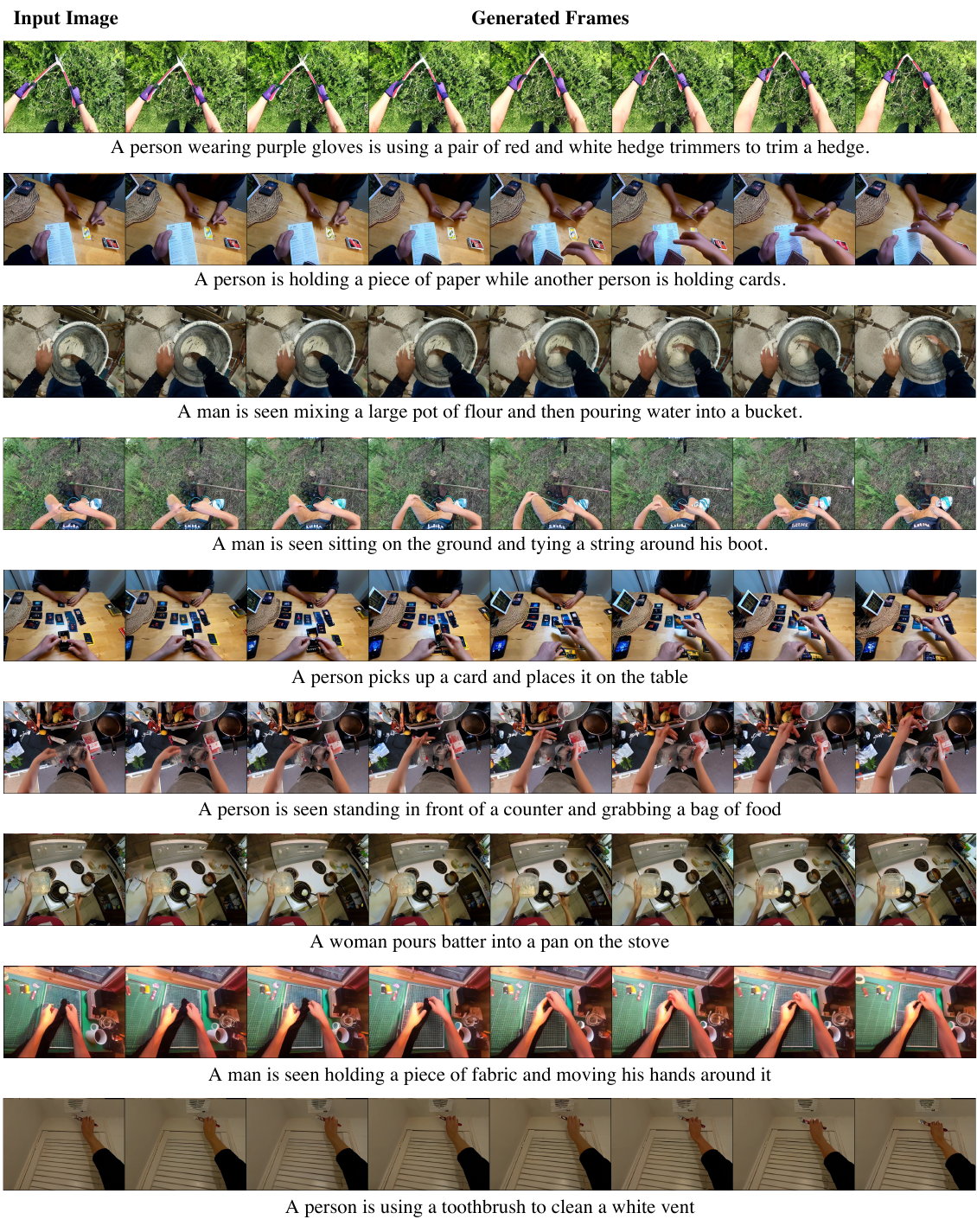}
    \caption{\textbf{Visualization of Video Generative Pretraining.} Given an input image and a text prompt, an I2V model is trained to predict the next 7 frames. Our pretrained video generation model is capable of generating plausible motions while maintaining the consistency with the input image.}
    \label{fig:vis_gen}
\end{figure*}

\subsection{Further Analysis}

\noindent\textbf{Visualization of Video Pretraining Model.}
The first stage of our proposed RynnVLA-001 involves pretraining an ego-centric Image-to-Video (I2V) model. This I2V paradigm is chosen to align with the typical input for VLA models: an initial image observation and a text-based instruction. As illustrated in Fig.~\ref{fig:vis_gen}, the pretrained model can generate video frames with plausible motion and consistent content from a given image and text prompt. Although the model is prone to generate subtle visual changes between frames, we find it sufficient for its role as a pretrained backbone for the subsequent VLA training.

\noindent\textbf{Improving Instruction-Following Capabilities by Increasing Scene Complexity in Training Data.}
Our evaluation protocol for instruction-following capabilities involves placing distractor objects in the desktop to test the model's robustness against visual ambiguity. We hypothesize that training exclusively on data with isolated target objects leads to a simplistic and vision-driven policy, where the model learns to grasp any objects without performing actions that follow the provided language instruction.
To validate this hypothesis, we perform an ablation study. A variant of RynnVLA-001 is trained solely on data without distractor objects. When evaluated on the task \texttt{pick up the strawberry} in a scene cluttered with pens and green blocks, this ablated model demonstrates a 0\% success rate over 10 trials. A total of 5 failure cases of the 10 trials consistently select a distractor object. In contrast, our full RynnVLA-001 model, trained on our comprehensive dataset including distractors, achieves a 90\% success rate (9/10) on this task. These results quantitatively underscore the critical importance of diverse data collection with distractors for developing reliable language-conditioned VLA models.

\begin{figure*}[t]
    \centering
    \includegraphics[width=1.0\textwidth]{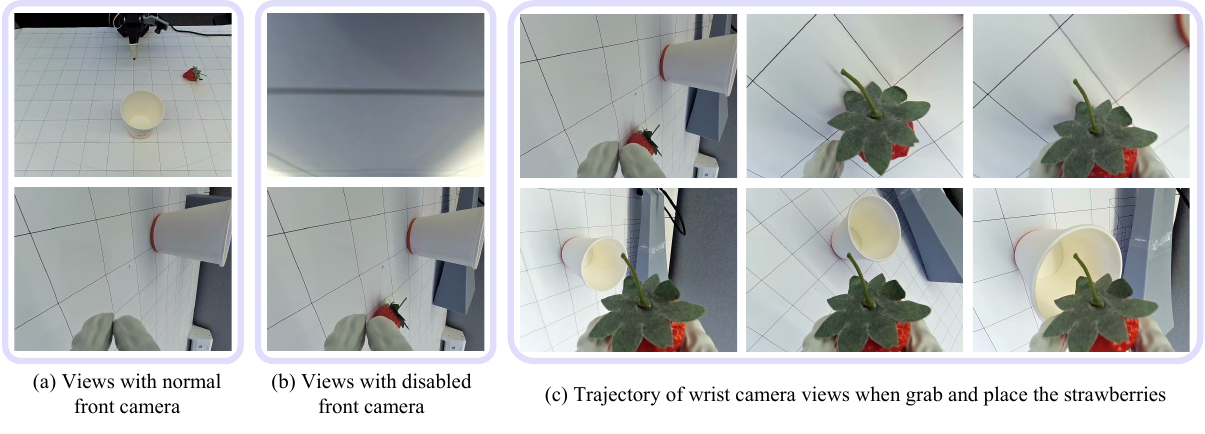}
    \caption{\textbf{Analysis on the front camera's function for coarse localization.} (a) Under normal dual-camera settings, the robot successfully picks the strawberries. (b) The front camera is masked, leaving only the wrist camera functional. (c) The robot can still complete the task if the target is within the wrist camera's initial field of view. However, task success rate drops from 80\% (4/5) to 0\% when the target is outside the wrist camera's view (on the left side), demonstrating that the front camera is essential for guiding the robot to the target's coarse location.}
    \label{fig:straw_disabled_front}
\end{figure*}

\begin{figure*}[t]
    \centering
    \includegraphics[width=1.0\textwidth]{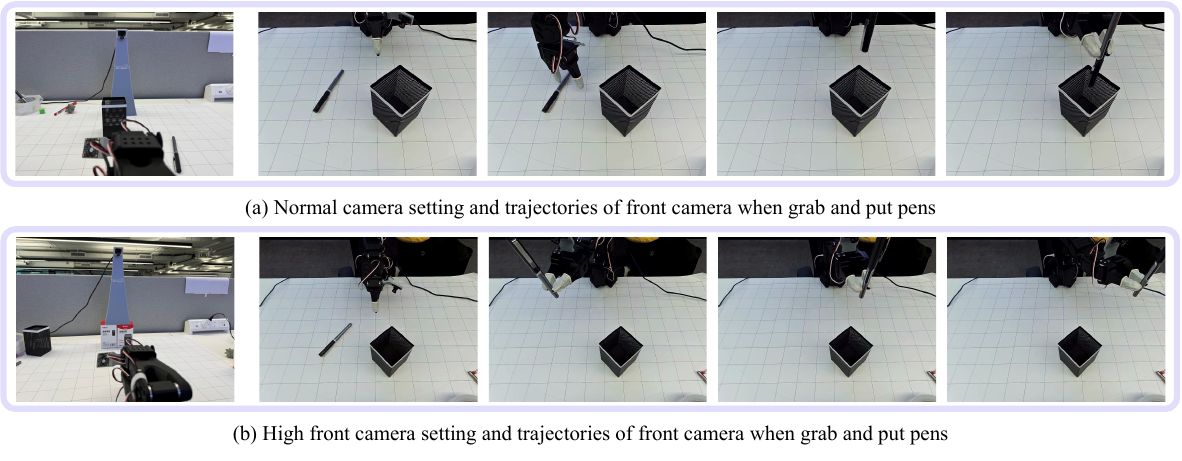}
    \caption{\textbf{The front camera provides critical 3D projective information for precise manipulation.} (a) With the standard camera configuration, the robot successfully inserts the pen into the holder. (b) When the front camera is elevated, the altered projective geometry of the scene causes the model to fail the task. This highlights the model's reliance on the specific 3D perspective provided by the front camera for spatial reasoning.}
    \label{fig:higher_front}
\end{figure*}

\noindent\textbf{Functional Analysis of Front and Wrist Cameras}
We also investigate the distinct functionalities of the front and wrist cameras on our LeRobot SO100 arm. We hypothesize that the front camera provides coarse object localization and 3D projective context, while the wrist camera is responsible for precise local adjustments.
1) To validate the front camera's role in coarse localization, we conduct an experiment where the front camera is disabled. Under normal conditions (Figure~\ref{fig:straw_disabled_front}(a)), the robot successfully completes the task. When the front camera is masked (Figure~\ref{fig:straw_disabled_front}(b)), we observe that the model could still succeed as long as the target is within the wrist camera's initial field of view (Figure~\ref{fig:straw_disabled_front}(c)). However, if the target (\eg, strawberries on the left) is outside the wrist camera's view, the robot fails to initiate any action. Quantitative results confirm this: for targets on the right, the success rate drops slightly from 100\% (5/5) to 80\% (4/5) after masking. For the targets on the left, the success rate decreases from 80\% (4/5) to 0\%. These findings strongly suggest that the front camera's primary function is to guide the end-effector to roughly approach the target.
2) Furthermore, we explore the front camera's function in providing 3D information for tasks requiring depth perception, such as inserting a pen into a holder. As shown in Figure~\ref{fig:higher_front}(a), the robot succeeds with the normal camera setup. However, when we elevate the front camera, altering the scene's projective geometry, the model fails to insert the pen (Figure~\ref{fig:higher_front}(b)). This demonstrates that the front camera provides critical 3D projective information that the model relies on for spatial reasoning and manipulation.

\section{Discussion and Conclusion}
\noindent

In this work, we propose RynnVLA-001, a VLA model enhanced by human demonstrations. We introduce human demonstrations in two pretraining stages: Ego-Centric Video Generative Pretraining and Human-Centric Trajectory-Aware Video Modeling. The first stage trains an I2V model by learning dynamics through predicting next frames. The second stage bridges the gaps between I2V models and VLA models by learning to predict keypoint trajectories of human. In VLA model, we propose ActionVAE to embed action chunks into a compacted embeddings. Owing to our dedicated designs, our proposed RynnVLA-001 outperforms state-of-the-art models such as GR00T N1.5 and Pi0.

\noindent\textbf{Limitation.} In this work, we validate the performance of RynnVLA-001 on the LeRobot SO100 robot arm. However, the scope of our current evaluation presents several limitations that we plan to address in future work. Our experiments are limited to a single robot embodiment and an evaluation environment that closely mirrored the training setup. Furthermore, the front-facing camera is mounted in a fixed position. To rigorously assess and enhance the model's generalization capabilities, future efforts will focus on: (1) extending the evaluation to a more diverse range of robot arms; (2) testing the model in more varied and unstructured environments; and (3) diversifying camera viewpoints.

\section*{Acknowledgement}
We thank our colleagues from the control team and the testing team for their valuable support throughout this work.
Our implementation is based on code from the open-source repositories \url{https://github.com/Alpha-VLLM/Lumina-mGPT} and \url{https://github.com/facebookresearch/chameleon}, which we adapted for training.
The LaTeX template is built upon Meta’s original template.

\bibliographystyle{assets/plainnat}
\bibliography{paper}

\begin{thebibliography}{77}
\providecommand{\natexlab}[1]{#1}
\providecommand{\url}[1]{\texttt{#1}}
\expandafter\ifx\csname urlstyle\endcsname\relax
  \providecommand{\doi}[1]{doi: #1}\else
  \providecommand{\doi}{doi: \begingroup \urlstyle{rm}\Url}\fi

\bibitem[Abadji et~al.(2022)Abadji, Suarez, Romary, and Sagot]{abadji2022towards}
Julien Abadji, Pedro~Ortiz Suarez, Laurent Romary, and Beno{\^\i}t Sagot.
\newblock Towards a cleaner document-oriented multilingual crawled corpus.
\newblock \emph{arXiv preprint arXiv:2201.06642}, 2022.

\bibitem[Anthropic()]{anthropic2025claude4}
Anthropic.
\newblock System card: {Claude Opus} 4 and {Claude Sonnet} 4.
\newblock \url{https://www.anthropic.com/news/claude-4}.

\bibitem[Assran et~al.(2023)Assran, Duval, Misra, Bojanowski, Vincent, Rabbat, LeCun, and Ballas]{assran2023ijepa}
Mahmoud Assran, Quentin Duval, Ishan Misra, Piotr Bojanowski, Pascal Vincent, Michael Rabbat, Yann LeCun, and Nicolas Ballas.
\newblock Self-supervised learning from images with a joint-embedding predictive architecture.
\newblock In \emph{IEEE/CVF Conference on Computer Vision and Pattern Recognition}, pages 15619--15629, 2023.

\bibitem[Bai et~al.(2025)Bai, Chen, Liu, Wang, Ge, Song, Dang, Wang, Wang, Tang, et~al.]{bai2025qwen2}
Shuai Bai, Keqin Chen, Xuejing Liu, Jialin Wang, Wenbin Ge, Sibo Song, Kai Dang, Peng Wang, Shijie Wang, Jun Tang, et~al.
\newblock {Qwen2.5-VL} technical report.
\newblock \emph{arXiv preprint arXiv:2502.13923}, 2025.

\bibitem[Baldridge et~al.(2024)Baldridge, Bauer, Bhutani, Brichtova, Bunner, Castrejon, Chan, Chen, Dieleman, Du, et~al.]{baldridge2024imagen}
Jason Baldridge, Jakob Bauer, Mukul Bhutani, Nicole Brichtova, Andrew Bunner, Lluis Castrejon, Kelvin Chan, Yichang Chen, Sander Dieleman, Yuqing Du, et~al.
\newblock Imagen 3.
\newblock \emph{arXiv preprint arXiv:2408.07009}, 2024.

\bibitem[Beyer et~al.(2024)Beyer, Steiner, Pinto, Kolesnikov, Wang, Salz, Neumann, Alabdulmohsin, Tschannen, Bugliarello, Unterthiner, Keysers, Koppula, Liu, Grycner, Gritsenko, Houlsby, Kumar, Rong, Eisenschlos, Kabra, Bauer, Bosnjak, Chen, Minderer, Voigtlaender, Bica, Balazevic, Puigcerver, Papalampidi, H{\'{e}}naff, Xiong, Soricut, Harmsen, and Zhai]{Beyer:PaliGemma}
Lucas Beyer, Andreas Steiner, Andr{\'{e}}~Susano Pinto, Alexander Kolesnikov, Xiao Wang, Daniel Salz, Maxim Neumann, Ibrahim Alabdulmohsin, Michael Tschannen, Emanuele Bugliarello, Thomas Unterthiner, Daniel Keysers, Skanda Koppula, Fangyu Liu, Adam Grycner, Alexey~A. Gritsenko, Neil Houlsby, Manoj Kumar, Keran Rong, Julian Eisenschlos, Rishabh Kabra, Matthias Bauer, Matko Bosnjak, Xi~Chen, Matthias Minderer, Paul Voigtlaender, Ioana Bica, Ivana Balazevic, Joan Puigcerver, Pinelopi Papalampidi, Olivier~J. H{\'{e}}naff, Xi~Xiong, Radu Soricut, Jeremiah Harmsen, and Xiaohua Zhai.
\newblock {PaliGemma:} {A} versatile {3B} {VLM} for transfer.
\newblock \emph{arXiv preprint arXiv:2407.07726}, 2024.

\bibitem[Bjorck et~al.(2025{\natexlab{a}})Bjorck, Blukis, Castañeda, Cherniadev, Da, Ding, Fan, Fang, Fox, Hu, Huang, Jang, Jiang, Kundalia, Kautz, Li, Lin, Lin, Magne, Man, Mandlekar, Narayan, Nasiriany, Reed, Tan, Wang, Wang, Wang, Wang, Xiang, Xie, Xu, Ye, Yu, Zhao, Zhang, Zheng, and Zhu]{Bjorck2024GR00T15}
Johan Bjorck, Valts Blukis, Fernando Castañeda, Nikita Cherniadev, Xingye Da, Runyu Ding, Linxi~"Jim" Fan, Yu~Fang, Dieter Fox, Fengyuan Hu, Spencer Huang, Joel Jang, Xiaowei Jiang, Kaushil Kundalia, Jan Kautz, Zhiqi Li, Kevin Lin, Zongyu Lin, Loic Magne, Yunze Man, Ajay Mandlekar, Avnish Narayan, Soroush Nasiriany, Scott Reed, You~Liang Tan, Guanzhi Wang, Jing Wang, Qi~Wang, Shihao Wang, Jiannan Xiang, Yuqi Xie, Yinzhen Xu, Seonghyeon Ye, Zhiding Yu, Yizhou Zhao, Zhe Zhang, Ruijie Zheng, and Yuke Zhu.
\newblock {GR00T N1.5}: An improved open foundation model for generalist humanoid robots.
\newblock \url{https://research.nvidia.com/labs/gear/gr00t-n1_5/}, 2025{\natexlab{a}}.

\bibitem[Bjorck et~al.(2025{\natexlab{b}})Bjorck, Casta{\~{n}}eda, Cherniadev, Da, Ding, Linxi, Fang, Fox, Hu, Huang, Jang, Jiang, Kautz, Kundalia, Lao, Li, Lin, Lin, Liu, LLontop, Magne, Mandlekar, Narayan, Nasiriany, Reed, Tan, Wang, Wang, Wang, Wang, Xiang, Xie, Xu, Xu, Ye, Yu, Zhang, Zhang, Zhao, Zheng, and Zhu]{Bjorck:GR00T-N1}
Johan Bjorck, Fernando Casta{\~{n}}eda, Nikita Cherniadev, Xingye Da, Runyu Ding, Linxi, Yu~Fang, Dieter Fox, Fengyuan Hu, Spencer Huang, Joel Jang, Zhenyu Jiang, Jan Kautz, Kaushil Kundalia, Lawrence Lao, Zhiqi Li, Zongyu Lin, Kevin Lin, Guilin Liu, Edith LLontop, Loic Magne, Ajay Mandlekar, Avnish Narayan, Soroush Nasiriany, Scott Reed, You~Liang Tan, Guanzhi Wang, Zu~Wang, Jing Wang, Qi~Wang, Jiannan Xiang, Yuqi Xie, Yinzhen Xu, Zhenjia Xu, Seonghyeon Ye, Zhiding Yu, Ao~Zhang, Hao Zhang, Yizhou Zhao, Ruijie Zheng, and Yuke Zhu.
\newblock {GR00T} {N1:} an open foundation model for generalist humanoid robots.
\newblock \emph{arXiv preprint arXiv:2503.14734}, 2025{\natexlab{b}}.

\bibitem[Black et~al.(2023)Black, Nakamoto, Atreya, Walke, Finn, Kumar, and Levine]{Black:SuSIE}
Kevin Black, Mitsuhiko Nakamoto, Pranav Atreya, Homer Walke, Chelsea Finn, Aviral Kumar, and Sergey Levine.
\newblock Zero-shot robotic manipulation with pretrained image-editing diffusion models.
\newblock \emph{arXiv preprint arXiv:2310.10639}, 2023.

\bibitem[Black et~al.(2024)Black, Brown, Driess, Esmail, Equi, Finn, Fusai, Groom, Hausman, Ichter, Jakubczak, Jones, Ke, Levine, Li{-}Bell, Mothukuri, Nair, Pertsch, Shi, Tanner, Vuong, Walling, Wang, and Zhilinsky]{Black:pi-0}
Kevin Black, Noah Brown, Danny Driess, Adnan Esmail, Michael Equi, Chelsea Finn, Niccolo Fusai, Lachy Groom, Karol Hausman, Brian Ichter, Szymon Jakubczak, Tim Jones, Liyiming Ke, Sergey Levine, Adrian Li{-}Bell, Mohith Mothukuri, Suraj Nair, Karl Pertsch, Lucy~Xiaoyang Shi, James Tanner, Quan Vuong, Anna Walling, Haohuan Wang, and Ury Zhilinsky.
\newblock {\(\pi\)}\({}_{\mbox{0}}\): {A} vision-language-action flow model for general robot control.
\newblock \emph{arXiv preprint arXiv:2410.24164}, 2024.

\bibitem[Brohan et~al.(2022)Brohan, Brown, Carbajal, Chebotar, Dabis, Finn, Gopalakrishnan, Hausman, Herzog, Hsu, et~al.]{brohan2022rt}
Anthony Brohan, Noah Brown, Justice Carbajal, Yevgen Chebotar, Joseph Dabis, Chelsea Finn, Keerthana Gopalakrishnan, Karol Hausman, Alex Herzog, Jasmine Hsu, et~al.
\newblock Rt-1: Robotics transformer for real-world control at scale.
\newblock \emph{arXiv preprint arXiv:2212.06817}, 2022.

\bibitem[Bu et~al.(2025)Bu, Cai, Chen, Cui, Ding, Feng, Gao, He, Hu, Huang, et~al.]{bu2025agibot}
Qingwen Bu, Jisong Cai, Li~Chen, Xiuqi Cui, Yan Ding, Siyuan Feng, Shenyuan Gao, Xindong He, Xuan Hu, Xu~Huang, et~al.
\newblock {AgiBot World Colosseo}: A large-scale manipulation platform for scalable and intelligent embodied systems.
\newblock \emph{arXiv preprint arXiv:2503.06669}, 2025.

\bibitem[Cadene et~al.(2024)Cadene, Alibert, Soare, Gallouedec, Zouitine, Palma, Kooijmans, Aractingi, Shukor, Aubakirova, Russi, Capuano, Pascal, Choghari, Moss, and Wolf]{cadene2024lerobot}
Remi Cadene, Simon Alibert, Alexander Soare, Quentin Gallouedec, Adil Zouitine, Steven Palma, Pepijn Kooijmans, Michel Aractingi, Mustafa Shukor, Dana Aubakirova, Martino Russi, Francesco Capuano, Caroline Pascal, Jade Choghari, Jess Moss, and Thomas Wolf.
\newblock {LeRobot}: State-of-the-art machine learning for real-world robotics in {Pytorch}.
\newblock \url{https://github.com/huggingface/lerobot}, 2024.

\bibitem[Cen et~al.(2025)Cen, Yu, Yuan, Jiang, Huang, Guo, Li, Song, Luo, Wang, Zhao, and Chen]{Cen:WorldVLA}
Jun Cen, Chaohui Yu, Hangjie Yuan, Yuming Jiang, Siteng Huang, Jiayan Guo, Xin Li, Yibing Song, Hao Luo, Fan Wang, Deli Zhao, and Hao Chen.
\newblock {WorldVLA}: Towards autoregressive action world model.
\newblock \emph{arXiv preprint arXiv:2506.21539}, 2025.

\bibitem[Cheang et~al.(2024)Cheang, Chen, Jing, Kong, Li, Li, Liu, Wu, Xu, Yang, Zhang, and Zhu]{Cheang:GR-2}
Chilam Cheang, Guangzeng Chen, Ya~Jing, Tao Kong, Hang Li, Yifeng Li, Yuxiao Liu, Hongtao Wu, Jiafeng Xu, Yichu Yang, Hanbo Zhang, and Minzhao Zhu.
\newblock {GR-2:} {A} generative video-language-action model with web-scale knowledge for robot manipulation.
\newblock \emph{arXiv preprint arXiv:2410.06158}, 2024.

\bibitem[Cheang et~al.(2025)Cheang, Chen, Cui, Hu, Huang, Kong, Li, Li, Liu, Ma, Niu, Ou, Peng, Ren, Shi, Tian, Wu, Xiao, Xiao, Xu, and Yang]{Cheang:GR-3}
Chilam Cheang, Sijin Chen, Zhongren Cui, Yingdong Hu, Liqun Huang, Tao Kong, Hang Li, Yifeng Li, Yuxiao Liu, Xiao Ma, Hao Niu, Wenxuan Ou, Wanli Peng, Zeyu Ren, Haixin Shi, Jiawen Tian, Hongtao Wu, Xin Xiao, Yuyang Xiao, Jiafeng Xu, and Yichu Yang.
\newblock {GR-3} technical report.
\newblock \emph{arXiv preprint arXiv:2507.15493}, 2025.

\bibitem[Chen et~al.(2024)Chen, Siarohin, Menapace, Deyneka, Chao, Jeon, Fang, Lee, Ren, Yang, et~al.]{chen2024panda}
Tsai-Shien Chen, Aliaksandr Siarohin, Willi Menapace, Ekaterina Deyneka, Hsiang-wei Chao, Byung~Eun Jeon, Yuwei Fang, Hsin-Ying Lee, Jian Ren, Ming-Hsuan Yang, et~al.
\newblock {Panda-70M}: Captioning {70M} videos with multiple cross-modality teachers.
\newblock In \emph{IEEE/CVF Conference on Computer Vision and Pattern Recognition}, pages 13320--13331, 2024.

\bibitem[Comanici et~al.(2025)Comanici, Bieber, Schaekermann, Pasupat, Sachdeva, Dhillon, Blistein, Ram, Zhang, Rosen, et~al.]{comanici2025gemini25}
Gheorghe Comanici, Eric Bieber, Mike Schaekermann, Ice Pasupat, Noveen Sachdeva, Inderjit Dhillon, Marcel Blistein, Ori Ram, Dan Zhang, Evan Rosen, et~al.
\newblock Gemini 2.5: Pushing the frontier with advanced reasoning, multimodality, long context, and next generation agentic capabilities.
\newblock \emph{arXiv preprint arXiv:2507.06261}, 2025.

\bibitem[Cui et~al.(2025)Cui, Ding, Song, Bai, Tong, Ge, Suo, Zhou, Liu, Jia, Zhao, Huang, and Wang]{Cui:Open-Helix}
Can Cui, Pengxiang Ding, Wenxuan Song, Shuanghao Bai, Xinyang Tong, Zirui Ge, Runze Suo, Wanqi Zhou, Yang Liu, Bofang Jia, Han Zhao, Siteng Huang, and Donglin Wang.
\newblock {OpenHelix:} {A} short survey, empirical analysis, and open-source dual-system {VLA} model for robotic manipulation.
\newblock \emph{arXiv preprint arXiv:2505.03912}, 2025.

\bibitem[Damen et~al.(2018)Damen, Doughty, Farinella, Fidler, Furnari, Kazakos, Moltisanti, Munro, Perrett, Price, and Wray]{Damen2018EPICKITCHENS}
Dima Damen, Hazel Doughty, Giovanni~Maria Farinella, Sanja Fidler, Antonino Furnari, Evangelos Kazakos, Davide Moltisanti, Jonathan Munro, Toby Perrett, Will Price, and Michael Wray.
\newblock Scaling egocentric vision: The {EPIC-KITCHENS} dataset.
\newblock In \emph{European Conference on Computer Vision}, 2018.

\bibitem[Damen et~al.(2021)Damen, Doughty, Farinella, Fidler, Furnari, Kazakos, Moltisanti, Munro, Perrett, Price, and Wray]{Damen2021PAMI}
Dima Damen, Hazel Doughty, Giovanni~Maria Farinella, Sanja Fidler, Antonino Furnari, Evangelos Kazakos, Davide Moltisanti, Jonathan Munro, Toby Perrett, Will Price, and Michael Wray.
\newblock The {EPIC-KITCHENS} dataset: Collection, challenges and baselines.
\newblock \emph{{IEEE} Transactions on Pattern Analysis and Machine Intelligence}, 43\penalty0 (11):\penalty0 4125--4141, 2021.

\bibitem[Damen et~al.(2022)Damen, Doughty, Farinella, Furnari, Ma, Kazakos, Moltisanti, Munro, Perrett, Price, and Wray]{Damen2022RESCALING}
Dima Damen, Hazel Doughty, Giovanni~Maria Farinella, Antonino Furnari, Jian Ma, Evangelos Kazakos, Davide Moltisanti, Jonathan Munro, Toby Perrett, Will Price, and Michael Wray.
\newblock Rescaling egocentric vision: Collection, pipeline and challenges for {EPIC-KITCHENS-100}.
\newblock \emph{International Journal of Computer Vision}, 130:\penalty0 33–55, 2022.

\bibitem[Du et~al.(2023)Du, Yang, Dai, Dai, Nachum, Tenenbaum, Schuurmans, and Abbeel]{Du:UniPi}
Yilun Du, Sherry Yang, Bo~Dai, Hanjun Dai, Ofir Nachum, Josh Tenenbaum, Dale Schuurmans, and Pieter Abbeel.
\newblock Learning universal policies via text-guided video generation.
\newblock In \emph{Advances in Neural Information Processing Systems}, 2023.

\bibitem[Esser et~al.(2024)Esser, Kulal, Blattmann, Entezari, M{\"u}ller, Saini, Levi, Lorenz, Sauer, Boesel, et~al.]{esser2024scaling}
Patrick Esser, Sumith Kulal, Andreas Blattmann, Rahim Entezari, Jonas M{\"u}ller, Harry Saini, Yam Levi, Dominik Lorenz, Axel Sauer, Frederic Boesel, et~al.
\newblock Scaling rectified flow transformers for high-resolution image synthesis.
\newblock In \emph{International Conference on Machine Learning}, 2024.

\bibitem[Goyal et~al.(2017)Goyal, Ebrahimi~Kahou, Michalski, Materzynska, Westphal, Kim, Haenel, Fruend, Yianilos, Mueller-Freitag, et~al.]{goyal2017something}
Raghav Goyal, Samira Ebrahimi~Kahou, Vincent Michalski, Joanna Materzynska, Susanne Westphal, Heuna Kim, Valentin Haenel, Ingo Fruend, Peter Yianilos, Moritz Mueller-Freitag, et~al.
\newblock The "{Something Something}" video database for learning and evaluating visual common sense.
\newblock In \emph{IEEE/CVF International Conference on Computer Vision}, pages 5842--5850, 2017.

\bibitem[Grattafiori et~al.(2024)Grattafiori, Dubey, Jauhri, Pandey, Kadian, Al-Dahle, Letman, Mathur, Schelten, Vaughan, et~al.]{grattafiori2024llama3}
Aaron Grattafiori, Abhimanyu Dubey, Abhinav Jauhri, Abhinav Pandey, Abhishek Kadian, Ahmad Al-Dahle, Aiesha Letman, Akhil Mathur, Alan Schelten, Alex Vaughan, et~al.
\newblock The llama 3 herd of models.
\newblock \emph{arXiv preprint arXiv:2407.21783}, 2024.

\bibitem[Grauman et~al.(2022)Grauman, Westbury, Byrne, Chavis, Furnari, Girdhar, Hamburger, Jiang, Liu, Liu, et~al.]{grauman2022ego4d}
Kristen Grauman, Andrew Westbury, Eugene Byrne, Zachary Chavis, Antonino Furnari, Rohit Girdhar, Jackson Hamburger, Hao Jiang, Miao Liu, Xingyu Liu, et~al.
\newblock {Ego4D}: Around the world in 3,000 hours of egocentric video.
\newblock In \emph{IEEE/CVF Conference on Computer Vision and Pattern Recognition}, pages 18995--19012, 2022.

\bibitem[Guo et~al.(2025{\natexlab{a}})Guo, Yang, Zhang, Song, Zhang, Xu, Zhu, Ma, Wang, Bi, et~al.]{guo2025deepseekr1}
Daya Guo, Dejian Yang, Haowei Zhang, Junxiao Song, Ruoyu Zhang, Runxin Xu, Qihao Zhu, Shirong Ma, Peiyi Wang, Xiao Bi, et~al.
\newblock {DeepSeek-R1}: Incentivizing reasoning capability in llms via reinforcement learning.
\newblock \emph{arXiv preprint arXiv:2501.12948}, 2025{\natexlab{a}}.

\bibitem[Guo et~al.(2025{\natexlab{b}})Guo, Wu, Zhu, Leng, Shi, Chen, Fan, Wang, Jiang, Wang, et~al.]{guo2025seed15}
Dong Guo, Faming Wu, Feida Zhu, Fuxing Leng, Guang Shi, Haobin Chen, Haoqi Fan, Jian Wang, Jianyu Jiang, Jiawei Wang, et~al.
\newblock {Seed1.5-VL} technical report.
\newblock \emph{arXiv preprint arXiv:2505.07062}, 2025{\natexlab{b}}.

\bibitem[Guo et~al.(2024)Guo, Hu, Zhang, Wang, Chen, Lu, and Chen]{Guo:PAD}
Yanjiang Guo, Yucheng Hu, Jianke Zhang, Yen{-}Jen Wang, Xiaoyu Chen, Chaochao Lu, and Jianyu Chen.
\newblock Prediction with action: Visual policy learning via joint denoising process.
\newblock In \emph{Advances in Neural Information Processing Systems}, 2024.

\bibitem[He et~al.(2022)He, Chen, Xie, Li, Doll{\'a}r, and Girshick]{he2022mae}
Kaiming He, Xinlei Chen, Saining Xie, Yanghao Li, Piotr Doll{\'a}r, and Ross Girshick.
\newblock Masked autoencoders are scalable vision learners.
\newblock In \emph{IEEE/CVF Conference on Computer Vision and Pattern Recognition}, pages 16000--16009, 2022.

\bibitem[Hoque et~al.(2025)Hoque, Huang, Yoon, Sivapurapu, and Zhang]{egodex_paper}
Ryan Hoque, Peide Huang, David~J. Yoon, Mouli Sivapurapu, and Jian Zhang.
\newblock {EgoDex}: Learning dexterous manipulation from large-scale egocentric video.
\newblock \emph{arXiv preprint arXiv:2505.11709}, 2025.

\bibitem[Hu et~al.(2024)Hu, Guo, Wang, Chen, Wang, Zhang, Sreenath, Lu, and Chen]{Hu:VPP}
Yucheng Hu, Yanjiang Guo, Pengchao Wang, Xiaoyu Chen, Yen{-}Jen Wang, Jianke Zhang, Koushil Sreenath, Chaochao Lu, and Jianyu Chen.
\newblock Video prediction policy: {A} generalist robot policy with predictive visual representations.
\newblock \emph{arXiv preprint arXiv:2412.14803}, 2024.

\bibitem[Jang et~al.(2025)Jang, Ye, Lin, Xiang, Bjorck, Fang, Hu, Huang, Kundalia, Lin, Magne, Mandlekar, Narayan, Tan, Wang, Wang, Wang, Xu, Zeng, Zheng, Zheng, Liu, Zettlemoyer, Fox, Kautz, Reed, Zhu, and Fan]{Jang2025DreamGenUG}
Joel Jang, Seonghyeon Ye, Zongyu Lin, Jiannan Xiang, Johan Bjorck, Yu~Fang, Fengyuan Hu, Spencer Huang, Kaushil Kundalia, Yen-Chen Lin, Loic Magne, Ajay Mandlekar, Avnish Narayan, You~Liang Tan, Guanzhi Wang, Jing Wang, Qi~Wang, Yinzhen Xu, Xi~Zeng, Kaiyuan Zheng, Ruijie Zheng, Ming-Yu Liu, Luke~S. Zettlemoyer, Dieter Fox, Jan Kautz, Scott Reed, Yuke Zhu, and Linxi Fan.
\newblock Dreamgen: Unlocking generalization in robot learning through neural trajectories.
\newblock \emph{arXiv preprint arXiv:2505.12705}, 2025.

\bibitem[Ke et~al.(2024)Ke, Gkanatsios, and Fragkiadaki]{Ke:3D-Diffuser-Actor}
Tsung{-}Wei Ke, Nikolaos Gkanatsios, and Katerina Fragkiadaki.
\newblock {3D Diffuser Actor:} policy diffusion with {3D} scene representations.
\newblock In \emph{Conference on Robot Learning}, pages 1949--1974, 2024.

\bibitem[Khazatsky et~al.(2024)Khazatsky, Pertsch, Nair, Balakrishna, Dasari, Karamcheti, Nasiriany, Srirama, Chen, Ellis, et~al.]{khazatsky2024droid}
Alexander Khazatsky, Karl Pertsch, Suraj Nair, Ashwin Balakrishna, Sudeep Dasari, Siddharth Karamcheti, Soroush Nasiriany, Mohan~Kumar Srirama, Lawrence~Yunliang Chen, Kirsty Ellis, et~al.
\newblock {DROID}: A large-scale in-the-wild robot manipulation dataset.
\newblock \emph{arXiv preprint arXiv:2403.12945}, 2024.

\bibitem[Kim et~al.(2024)Kim, Pertsch, Karamcheti, Xiao, Balakrishna, Nair, Rafailov, Foster, Sanketi, Vuong, Kollar, Burchfiel, Tedrake, Sadigh, Levine, Liang, and Finn]{Kim:OpenVLA}
Moo~Jin Kim, Karl Pertsch, Siddharth Karamcheti, Ted Xiao, Ashwin Balakrishna, Suraj Nair, Rafael Rafailov, Ethan~Paul Foster, Pannag~R. Sanketi, Quan Vuong, Thomas Kollar, Benjamin Burchfiel, Russ Tedrake, Dorsa Sadigh, Sergey Levine, Percy Liang, and Chelsea Finn.
\newblock {OpenVLA:} an open-source vision-language-action model.
\newblock In \emph{Conference on Robot Learning}, pages 2679--2713, 2024.

\bibitem[Kim et~al.(2025)Kim, Finn, and Liang]{kim2025fine}
Moo~Jin Kim, Chelsea Finn, and Percy Liang.
\newblock Fine-tuning vision-language-action models: Optimizing speed and success.
\newblock \emph{arXiv preprint arXiv:2502.19645}, 2025.

\bibitem[Li et~al.(2024{\natexlab{a}})Li, Fang, Smyrnis, Ivgi, Jordan, Gadre, Bansal, Guha, Keh, Arora, et~al.]{li2024datacomp}
Jeffrey Li, Alex Fang, Georgios Smyrnis, Maor Ivgi, Matt Jordan, Samir~Yitzhak Gadre, Hritik Bansal, Etash Guha, Sedrick~Scott Keh, Kushal Arora, et~al.
\newblock {DataComp-LM}: In search of the next generation of training sets for language models.
\newblock \emph{Advances in Neural Information Processing Systems}, 37:\penalty0 14200--14282, 2024{\natexlab{a}}.

\bibitem[Li et~al.(2024{\natexlab{b}})Li, Liang, Wang, Luo, Chen, Liao, Wei, Deng, Xu, Zhang, Wang, Liu, Fu, Bao, Chen, Shi, Yang, and Guo]{Li:CogACT}
Qixiu Li, Yaobo Liang, Zeyu Wang, Lin Luo, Xi~Chen, Mozheng Liao, Fangyun Wei, Yu~Deng, Sicheng Xu, Yizhong Zhang, Xiaofan Wang, Bei Liu, Jianlong Fu, Jianmin Bao, Dong Chen, Yuanchun Shi, Jiaolong Yang, and Baining Guo.
\newblock {CogACT:} {A} foundational vision-language-action model for synergizing cognition and action in robotic manipulation.
\newblock \emph{arXiv preprint arXiv:2411.19650}, 2024{\natexlab{b}}.

\bibitem[Lin et~al.(2025)Lin, Nai, Hu, You, Zhao, and Gao]{Lin:OneTwoVLA}
Fanqi Lin, Ruiqian Nai, Yingdong Hu, Jiacheng You, Junming Zhao, and Yang Gao.
\newblock {OneTwoVLA}: {A} unified vision-language-action model with adaptive reasoning.
\newblock \emph{arXiv preprint arXiv:2505.11917}, 2025.

\bibitem[Lipman et~al.(2023)Lipman, Chen, Ben{-}Hamu, Nickel, and Le]{Lipman:flow-matching}
Yaron Lipman, Ricky T.~Q. Chen, Heli Ben{-}Hamu, Maximilian Nickel, and Matthew Le.
\newblock Flow matching for generative modeling.
\newblock In \emph{International Conference on Learning Representations}, 2023.

\bibitem[Liu et~al.(2024{\natexlab{a}})Liu, Zhao, Zhuo, Lin, Qiao, Li, and Gao]{lumina-mgpt}
Dongyang Liu, Shitian Zhao, Le~Zhuo, Weifeng Lin, Yu~Qiao, Hongsheng Li, and Peng Gao.
\newblock {Lumina-mGPT}: Illuminate flexible photorealistic text-to-image generation with multimodal generative pretraining.
\newblock \emph{arXiv preprint arXiv:2408.02657}, 2024{\natexlab{a}}.

\bibitem[Liu et~al.(2023)Liu, Li, Wu, and Lee]{Liu:LLaVA}
Haotian Liu, Chunyuan Li, Qingyang Wu, and Yong~Jae Lee.
\newblock Visual instruction tuning.
\newblock In \emph{Advances in Neural Information Processing Systems}, pages 34892--34916, 2023.

\bibitem[Liu et~al.(2024{\natexlab{b}})Liu, Wu, Li, Tan, Chen, Wang, Xu, Su, and Zhu]{liu2024rdt1b}
Songming Liu, Lingxuan Wu, Bangguo Li, Hengkai Tan, Huayu Chen, Zhengyi Wang, Ke~Xu, Hang Su, and Jun Zhu.
\newblock {RDT-1B}: a diffusion foundation model for bimanual manipulation.
\newblock \emph{arXiv preprint arXiv:2410.07864}, 2024{\natexlab{b}}.

\bibitem[Mahdisoltani et~al.(2018)Mahdisoltani, Berger, Gharbieh, Fleet, and Memisevic]{mahdisoltani2018effectiveness}
Farzaneh Mahdisoltani, Guillaume Berger, Waseem Gharbieh, David Fleet, and Roland Memisevic.
\newblock On the effectiveness of task granularity for transfer learning.
\newblock \emph{arXiv preprint arXiv:1804.09235}, 2018.

\bibitem[Mees et~al.(2022)Mees, Hermann, Rosete-Beas, and Burgard]{mees2022calvin}
Oier Mees, Lukas Hermann, Erick Rosete-Beas, and Wolfram Burgard.
\newblock {CALVIN}: A benchmark for language-conditioned policy learning for long-horizon robot manipulation tasks.
\newblock \emph{IEEE Robotics and Automation Letters}, 7\penalty0 (3):\penalty0 7327--7334, 2022.

\bibitem[Miech et~al.(2019)Miech, Zhukov, Alayrac, Tapaswi, Laptev, and Sivic]{miech19howto100m}
Antoine Miech, Dimitri Zhukov, Jean-Baptiste Alayrac, Makarand Tapaswi, Ivan Laptev, and Josef Sivic.
\newblock How{T}o100{M}: {L}earning a {T}ext-{V}ideo {E}mbedding by {W}atching {H}undred {M}illion {N}arrated {V}ideo {C}lips.
\newblock In \emph{IEEE/CVF International Conference on Computer Vision}, 2019.

\bibitem[O'Neill et~al.(2024)O'Neill, Rehman, Maddukuri, Gupta, Padalkar, Lee, Pooley, Gupta, Mandlekar, Jain, Tung, Bewley, Herzog, Irpan, Khazatsky, Rai, Gupta, Wang, Singh, Garg, Kembhavi, Xie, Brohan, Raffin, Sharma, Yavary, Jain, Balakrishna, Wahid, Burgess{-}Limerick, Kim, Sch{\"{o}}lkopf, Wulfe, Ichter, Lu, Xu, Le, Finn, Wang, Xu, Chi, Huang, Chan, Agia, Pan, Fu, Devin, Xu, Morton, Driess, Chen, Pathak, Shah, B{\"{u}}chler, Jayaraman, Kalashnikov, Sadigh, Johns, Foster, Liu, Ceola, Xia, Zhao, Stulp, Zhou, Sukhatme, Salhotra, Yan, Feng, Schiavi, Berseth, Kahn, Wang, Su, Fang, Shi, Bao, Amor, Christensen, Furuta, Walke, Fang, Ha, Mordatch, Radosavovic, Leal, Liang, Abou{-}Chakra, Kim, Drake, Peters, Schneider, Hsu, Bohg, Bingham, Wu, Gao, Hu, Wu, Wu, Sun, Luo, Gu, Tan, Oh, Wu, Lu, Yang, Malik, Silv{\'{e}}rio, Hejna, Booher, Tompson, Yang, Salvador, Lim, Han, Wang, Rao, Pertsch, Hausman, Go, Gopalakrishnan, Goldberg, Byrne, Oslund, Kawaharazuka, Black, Lin, Zhang, Ehsani, Lekkala, Ellis, Rana, Srinivasan,
  Fang, Singh, Zeng, Hatch, Hsu, Itti, Chen, Pinto, Fei{-}Fei, Tan, Fan, Ott, Lee, Weihs, Chen, Lepert, Memmel, Tomizuka, Itkina, Castro, Spero, Du, Ahn, Yip, Zhang, Ding, Heo, Srirama, Sharma, Kim, Kanazawa, Hansen, Heess, Joshi, S{\"{u}}nderhauf, Liu, Palo, Shafiullah, Mees, Kroemer, Bastani, Sanketi, Miller, Yin, Wohlhart, Xu, Fagan, Mitrano, Sermanet, Abbeel, Sundaresan, Chen, Vuong, Rafailov, Tian, Doshi, Mart{\'{\i}}n{-}Mart{\'{\i}}n, Baijal, Scalise, Hendrix, Lin, Qian, Zhang, Mendonca, Shah, Hoque, Julian, Bustamante, Kirmani, Levine, Lin, Moore, Bahl, Dass, Sonawani, Song, Xu, Haldar, Karamcheti, Adebola, Guist, Nasiriany, Schaal, Welker, Tian, Ramamoorthy, Dasari, Belkhale, Park, Nair, Mirchandani, Osa, Gupta, Harada, Matsushima, Xiao, Kollar, Yu, Ding, Davchev, Zhao, Armstrong, Darrell, Chung, Jain, Vanhoucke, Zhan, Zhou, Burgard, Chen, Wang, Zhu, Geng, Liu, Xu, Li, Lu, Ma, Kim, Chebotar, Zhou, Zhu, Wu, Xu, Wang, Bisk, Cho, Lee, Cui, Cao, Wu, Tang, Zhu, Zhang, Jiang, Li, Li, Iwasawa, Matsuo, Ma,
  Xu, Cui, Zhang, and Lin]{O'Neill:OXE}
Abby O'Neill, Abdul Rehman, Abhiram Maddukuri, Abhishek Gupta, Abhishek Padalkar, Abraham Lee, Acorn Pooley, Agrim Gupta, Ajay Mandlekar, Ajinkya Jain, Albert Tung, Alex Bewley, Alexander Herzog, Alex Irpan, Alexander Khazatsky, Anant Rai, Anchit Gupta, Andrew~E. Wang, Anikait Singh, Animesh Garg, Aniruddha Kembhavi, Annie Xie, Anthony Brohan, Antonin Raffin, Archit Sharma, Arefeh Yavary, Arhan Jain, Ashwin Balakrishna, Ayzaan Wahid, Ben Burgess{-}Limerick, Beomjoon Kim, Bernhard Sch{\"{o}}lkopf, Blake Wulfe, Brian Ichter, Cewu Lu, Charles Xu, Charlotte Le, Chelsea Finn, Chen Wang, Chenfeng Xu, Cheng Chi, Chenguang Huang, Christine Chan, Christopher Agia, Chuer Pan, Chuyuan Fu, Coline Devin, Danfei Xu, Daniel Morton, Danny Driess, Daphne Chen, Deepak Pathak, Dhruv Shah, Dieter B{\"{u}}chler, Dinesh Jayaraman, Dmitry Kalashnikov, Dorsa Sadigh, Edward Johns, Ethan~Paul Foster, Fangchen Liu, Federico Ceola, Fei Xia, Feiyu Zhao, Freek Stulp, Gaoyue Zhou, Gaurav~S. Sukhatme, Gautam Salhotra, Ge~Yan, Gilbert Feng,
  Giulio Schiavi, Glen Berseth, Gregory Kahn, Guanzhi Wang, Hao Su, Haoshu Fang, Haochen Shi, Henghui Bao, Heni~Ben Amor, Henrik~I. Christensen, Hiroki Furuta, Homer Walke, Hongjie Fang, Huy Ha, Igor Mordatch, Ilija Radosavovic, Isabel Leal, Jacky Liang, Jad Abou{-}Chakra, Jaehyung Kim, Jaimyn Drake, Jan Peters, Jan Schneider, Jasmine Hsu, Jeannette Bohg, Jeffrey Bingham, Jeffrey Wu, Jensen Gao, Jiaheng Hu, Jiajun Wu, Jialin Wu, Jiankai Sun, Jianlan Luo, Jiayuan Gu, Jie Tan, Jihoon Oh, Jimmy Wu, Jingpei Lu, Jingyun Yang, Jitendra Malik, Jo{\~{a}}o Silv{\'{e}}rio, Joey Hejna, Jonathan Booher, Jonathan Tompson, Jonathan Yang, Jordi Salvador, Joseph~J. Lim, Junhyek Han, Kaiyuan Wang, Kanishka Rao, Karl Pertsch, Karol Hausman, Keegan Go, Keerthana Gopalakrishnan, Ken Goldberg, Kendra Byrne, Kenneth Oslund, Kento Kawaharazuka, Kevin Black, Kevin Lin, Kevin Zhang, Kiana Ehsani, Kiran Lekkala, Kirsty Ellis, Krishan Rana, Krishnan Srinivasan, Kuan Fang, Kunal~Pratap Singh, Kuo{-}Hao Zeng, Kyle Hatch, Kyle Hsu,
  Laurent Itti, Lawrence~Yunliang Chen, Lerrel Pinto, Li~Fei{-}Fei, Liam Tan, Linxi~Jim Fan, Lionel Ott, Lisa Lee, Luca Weihs, Magnum Chen, Marion Lepert, Marius Memmel, Masayoshi Tomizuka, Masha Itkina, Mateo~Guaman Castro, Max Spero, Maximilian Du, Michael Ahn, Michael~C. Yip, Mingtong Zhang, Mingyu Ding, Minho Heo, Mohan~Kumar Srirama, Mohit Sharma, Moo~Jin Kim, Naoaki Kanazawa, Nicklas Hansen, Nicolas Heess, Nikhil~J. Joshi, Niko S{\"{u}}nderhauf, Ning Liu, Norman~Di Palo, Nur Muhammad~(Mahi) Shafiullah, Oier Mees, Oliver Kroemer, Osbert Bastani, Pannag~R. Sanketi, Patrick~Tree Miller, Patrick Yin, Paul Wohlhart, Peng Xu, Peter~David Fagan, Peter Mitrano, Pierre Sermanet, Pieter Abbeel, Priya Sundaresan, Qiuyu Chen, Quan Vuong, Rafael Rafailov, Ran Tian, Ria Doshi, Roberto Mart{\'{\i}}n{-}Mart{\'{\i}}n, Rohan Baijal, Rosario Scalise, Rose Hendrix, Roy Lin, Runjia Qian, Ruohan Zhang, Russell Mendonca, Rutav Shah, Ryan Hoque, Ryan Julian, Samuel Bustamante, Sean Kirmani, Sergey Levine, Shan Lin, Sherry
  Moore, Shikhar Bahl, Shivin Dass, Shubham~D. Sonawani, Shuran Song, Sichun Xu, Siddhant Haldar, Siddharth Karamcheti, Simeon Adebola, Simon Guist, Soroush Nasiriany, Stefan Schaal, Stefan Welker, Stephen Tian, Subramanian Ramamoorthy, Sudeep Dasari, Suneel Belkhale, Sungjae Park, Suraj Nair, Suvir Mirchandani, Takayuki Osa, Tanmay Gupta, Tatsuya Harada, Tatsuya Matsushima, Ted Xiao, Thomas Kollar, Tianhe Yu, Tianli Ding, Todor Davchev, Tony~Z. Zhao, Travis Armstrong, Trevor Darrell, Trinity Chung, Vidhi Jain, Vincent Vanhoucke, Wei Zhan, Wenxuan Zhou, Wolfram Burgard, Xi~Chen, Xiaolong Wang, Xinghao Zhu, Xinyang Geng, Xiyuan Liu, Liangwei Xu, Xuanlin Li, Yao Lu, Yecheng~Jason Ma, Yejin Kim, Yevgen Chebotar, Yifan Zhou, Yifeng Zhu, Yilin Wu, Ying Xu, Yixuan Wang, Yonatan Bisk, Yoonyoung Cho, Youngwoon Lee, Yuchen Cui, Yue Cao, Yueh{-}Hua Wu, Yujin Tang, Yuke Zhu, Yunchu Zhang, Yunfan Jiang, Yunshuang Li, Yunzhu Li, Yusuke Iwasawa, Yutaka Matsuo, Zehan Ma, Zhuo Xu, Zichen~Jeff Cui, Zichen Zhang, and Zipeng
  Lin.
\newblock {Open X-Embodiment:} robotic learning datasets and {RT-X} models : {Open X-Embodiment} collaboration.
\newblock In \emph{IEEE International Conference on Robotics and Automation}, pages 6892--6903, 2024.

\bibitem[OpenAI(2024)]{openai2024gpt4o}
OpenAI.
\newblock {GPT-4o} system card, 2024.
\newblock \url{https://openai.com/index/hello-gpt-4o/}.

\bibitem[OpenAI(2025)]{openai2025gpt41}
OpenAI.
\newblock Introducing {GPT-4.1} in the {API}, 2025.
\newblock \url{https://openai.com/index/gpt-4-1/}.

\bibitem[Oquab et~al.(2023)Oquab, Darcet, Moutakanni, Vo, Szafraniec, Khalidov, Fernandez, Haziza, Massa, El-Nouby, et~al.]{oquab2023dinov2}
Maxime Oquab, Timoth{\'e}e Darcet, Th{\'e}o Moutakanni, Huy Vo, Marc Szafraniec, Vasil Khalidov, Pierre Fernandez, Daniel Haziza, Francisco Massa, Alaaeldin El-Nouby, et~al.
\newblock {DINOv2}: Learning robust visual features without supervision.
\newblock \emph{arXiv preprint arXiv:2304.07193}, 2023.

\bibitem[Peebles and Xie(2023)]{Peebles:DiT}
William Peebles and Saining Xie.
\newblock Scalable diffusion models with transformers.
\newblock In \emph{IEEE/CVF International Conference on Computer Vision}, pages 4172--4182, 2023.

\bibitem[Pertsch et~al.(2025)Pertsch, Stachowicz, Ichter, Driess, Nair, Vuong, Mees, Finn, and Levine]{Pertsch:FAST}
Karl Pertsch, Kyle Stachowicz, Brian Ichter, Danny Driess, Suraj Nair, Quan Vuong, Oier Mees, Chelsea Finn, and Sergey Levine.
\newblock {FAST:} efficient action tokenization for vision-language-action models.
\newblock \emph{arXiv preprint arXiv:2501.09747}, 2025.

\bibitem[Ravi et~al.(2024)Ravi, Gabeur, Hu, Hu, Ryali, Ma, Khedr, R{\"a}dle, Rolland, Gustafson, et~al.]{ravi2024sam2}
Nikhila Ravi, Valentin Gabeur, Yuan-Ting Hu, Ronghang Hu, Chaitanya Ryali, Tengyu Ma, Haitham Khedr, Roman R{\"a}dle, Chloe Rolland, Laura Gustafson, et~al.
\newblock {SAM} 2: Segment anything in images and videos.
\newblock \emph{arXiv preprint arXiv:2408.00714}, 2024.

\bibitem[Schuhmann et~al.(2022)Schuhmann, Beaumont, Vencu, Gordon, Wightman, Cherti, Coombes, Katta, Mullis, Wortsman, et~al.]{schuhmann2022laion}
Christoph Schuhmann, Romain Beaumont, Richard Vencu, Cade Gordon, Ross Wightman, Mehdi Cherti, Theo Coombes, Aarush Katta, Clayton Mullis, Mitchell Wortsman, et~al.
\newblock {LAION-5B}: An open large-scale dataset for training next generation image-text models.
\newblock \emph{Advances in Neural Information Processing Systems}, pages 25278--25294, 2022.

\bibitem[Shentu et~al.(2024)Shentu, Wu, Rajeswaran, and Abbeel]{Shentu:LCB}
Yide Shentu, Philipp Wu, Aravind Rajeswaran, and Pieter Abbeel.
\newblock From {LLMs} to actions: Latent codes as bridges in hierarchical robot control.
\newblock In \emph{IEEE/RSJ International Conference on Intelligent Robots and Systems}, pages 8539--8546, 2024.

\bibitem[Shukor et~al.(2025)Shukor, Aubakirova, Capuano, Kooijmans, Palma, Zouitine, Aractingi, Pascal, Russi, Marafioti, Alibert, Cord, Wolf, and Cad{\`{e}}ne]{Shukor:SmolVLA}
Mustafa Shukor, Dana Aubakirova, Francesco Capuano, Pepijn Kooijmans, Steven Palma, Adil Zouitine, Michel Aractingi, Caroline Pascal, Martino Russi, Andr{\'{e}}s Marafioti, Simon Alibert, Matthieu Cord, Thomas Wolf, and R{\'{e}}mi Cad{\`{e}}ne.
\newblock {SmolVLA:} {A} vision-language-action model for affordable and efficient robotics.
\newblock \emph{arXiv preprint arXiv:2506.01844}, 2025.

\bibitem[Team(2024)]{Chameleonpaper}
Chameleon Team.
\newblock Chameleon: Mixed-modal early-fusion foundation models.
\newblock \emph{arXiv preprint arXiv:2405.09818}, 2024.

\bibitem[Tian et~al.(2024)Tian, Jiang, Yuan, Peng, and Wang]{tian2024visual}
Keyu Tian, Yi~Jiang, Zehuan Yuan, Bingyue Peng, and Liwei Wang.
\newblock Visual autoregressive modeling: Scalable image generation via next-scale prediction.
\newblock In \emph{Advances in Neural Information Processing Systems}, pages 84839--84865, 2024.

\bibitem[Tian et~al.(2025)Tian, Yang, Zeng, Wang, Lin, Dong, and Pang]{Tian:Seer}
Yang Tian, Sizhe Yang, Jia Zeng, Ping Wang, Dahua Lin, Hao Dong, and Jiangmiao Pang.
\newblock Predictive inverse dynamics models are scalable learners for robotic manipulation.
\newblock In \emph{International Conference on Learning Representations}, 2025.

\bibitem[Tschannen et~al.(2025)Tschannen, Gritsenko, Wang, Naeem, Alabdulmohsin, Parthasarathy, Evans, Beyer, Xia, Mustafa, et~al.]{tschannen2025siglip2}
Michael Tschannen, Alexey Gritsenko, Xiao Wang, Muhammad~Ferjad Naeem, Ibrahim Alabdulmohsin, Nikhil Parthasarathy, Talfan Evans, Lucas Beyer, Ye~Xia, Basil Mustafa, et~al.
\newblock {SigLIP} 2: Multilingual vision-language encoders with improved semantic understanding, localization, and dense features.
\newblock \emph{arXiv preprint arXiv:2502.14786}, 2025.

\bibitem[Walke et~al.(2023)Walke, Black, Lee, Kim, Du, Zheng, Zhao, Hansen-Estruch, Vuong, He, Myers, Fang, Finn, and Levine]{walke2023bridgedata}
Homer Walke, Kevin Black, Abraham Lee, Moo~Jin Kim, Max Du, Chongyi Zheng, Tony Zhao, Philippe Hansen-Estruch, Quan Vuong, Andre He, Vivek Myers, Kuan Fang, Chelsea Finn, and Sergey Levine.
\newblock {BridgeData V2}: A dataset for robot learning at scale.
\newblock In \emph{Conference on Robot Learning}, 2023.

\bibitem[Wang et~al.(2024{\natexlab{a}})Wang, Bai, Tan, Wang, Fan, Bai, Chen, Liu, Wang, Ge, Fan, Dang, Du, Ren, Men, Liu, Zhou, Zhou, and Lin]{Qwen2VL}
Peng Wang, Shuai Bai, Sinan Tan, Shijie Wang, Zhihao Fan, Jinze Bai, Keqin Chen, Xuejing Liu, Jialin Wang, Wenbin Ge, Yang Fan, Kai Dang, Mengfei Du, Xuancheng Ren, Rui Men, Dayiheng Liu, Chang Zhou, Jingren Zhou, and Junyang Lin.
\newblock {Qwen2-VL}: Enhancing vision-language model's perception of the world at any resolution.
\newblock \emph{arXiv preprint arXiv:2409.12191}, 2024{\natexlab{a}}.

\bibitem[Wang et~al.(2024{\natexlab{b}})Wang, Zhao, Liu, Wang, Zhao, Bao, Zhu, Zhang, and Wang]{wang2024egovid}
Xiaofeng Wang, Kang Zhao, Feng Liu, Jiayu Wang, Guosheng Zhao, Xiaoyi Bao, Zheng Zhu, Yingya Zhang, and Xingang Wang.
\newblock {EgoVid-5M}: A large-scale video-action dataset for egocentric video generation.
\newblock \emph{arXiv preprint arXiv:2411.08380}, 2024{\natexlab{b}}.

\bibitem[Wang et~al.(2025)Wang, Zhu, Liu, Yang, Fang, and He]{wang2025vq}
Yating Wang, Haoyi Zhu, Mingyu Liu, Jiange Yang, Hao-Shu Fang, and Tong He.
\newblock {VQ-VLA}: Improving vision-language-action models via scaling vector-quantized action tokenizers.
\newblock \emph{arXiv preprint arXiv:2507.01016}, 2025.

\bibitem[Wen et~al.(2025)Wen, Zhu, Li, Tang, Shen, and Feng]{Wen:DexVLA}
Junjie Wen, Yichen Zhu, Jinming Li, Zhibin Tang, Chaomin Shen, and Feifei Feng.
\newblock {DexVLA:} vision-language model with plug-in diffusion expert for general robot control.
\newblock \emph{arXiv preprint arXiv:2502.05855}, 2025.

\bibitem[Wu et~al.(2024)Wu, Jing, Cheang, Chen, Xu, Li, Liu, Li, and Kong]{Wu:GR-1}
Hongtao Wu, Ya~Jing, Chilam Cheang, Guangzeng Chen, Jiafeng Xu, Xinghang Li, Minghuan Liu, Hang Li, and Tao Kong.
\newblock Unleashing large-scale video generative pre-training for visual robot manipulation.
\newblock In \emph{International Conference on Learning Representations}, 2024.

\bibitem[Yang et~al.(2025)Yang, Li, Yang, Zhang, Hui, Zheng, Yu, Gao, Huang, Lv, et~al.]{yang2025qwen3}
An~Yang, Anfeng Li, Baosong Yang, Beichen Zhang, Binyuan Hui, Bo~Zheng, Bowen Yu, Chang Gao, Chengen Huang, Chenxu Lv, et~al.
\newblock Qwen3 technical report.
\newblock \emph{arXiv preprint arXiv:2505.09388}, 2025.

\bibitem[Yang et~al.(2023)Yang, Zeng, Yuan, and Li]{yang2023effective}
Zhendong Yang, Ailing Zeng, Chun Yuan, and Yu~Li.
\newblock Effective whole-body pose estimation with two-stages distillation.
\newblock In \emph{IEEE/CVF Conference on Computer Vision and Pattern Recognition}, pages 4210--4220, 2023.

\bibitem[Zhang et~al.(2025)Zhang, Ding, Lyu, Peng, and Wang]{Zhang:GEVRM}
Hongyin Zhang, Pengxiang Ding, Shangke Lyu, Ying Peng, and Donglin Wang.
\newblock {GEVRM:} goal-expressive video generation model for robust visual manipulation.
\newblock In \emph{International Conference on Learning Representations}, 2025.

\bibitem[Zhang et~al.(2024)Zhang, Guo, Chen, Wang, Hu, Shi, and Chen]{Zhang:HiRT}
Jianke Zhang, Yanjiang Guo, Xiaoyu Chen, Yen{-}Jen Wang, Yucheng Hu, Chengming Shi, and Jianyu Chen.
\newblock {HiRT:} enhancing robotic control with hierarchical robot transformers.
\newblock In \emph{Conference on Robot Learning}, pages 933--946, 2024.

\bibitem[Zhao et~al.(2025)Zhao, Lu, Kim, Fu, Zhang, Wu, Li, Ma, Han, Finn, Handa, Lin, Wetzstein, Liu, and Xiang]{Zhao:CoT-VLA}
Qingqing Zhao, Yao Lu, Moo~Jin Kim, Zipeng Fu, Zhuoyang Zhang, Yecheng Wu, Zhaoshuo Li, Qianli Ma, Song Han, Chelsea Finn, Ankur Handa, Tsung{-}Yi Lin, Gordon Wetzstein, Ming{-}Yu Liu, and Donglai Xiang.
\newblock {CoT-VLA}: Visual chain-of-thought reasoning for vision-language-action models.
\newblock In \emph{IEEE/CVF Conference on Computer Vision and Pattern Recognition}, pages 1702--1713, 2025.

\bibitem[Zhao et~al.(2023)Zhao, Kumar, Levine, and Finn]{zhao2023learning}
Tony~Z Zhao, Vikash Kumar, Sergey Levine, and Chelsea Finn.
\newblock Learning fine-grained bimanual manipulation with low-cost hardware.
\newblock \emph{arXiv preprint arXiv:2304.13705}, 2023.

\bibitem[Zheng et~al.(2024)Zheng, Peng, Yang, Shen, Li, Liu, Zhou, Li, and You]{Zheng:Open-Sora}
Zangwei Zheng, Xiangyu Peng, Tianji Yang, Chenhui Shen, Shenggui Li, Hongxin Liu, Yukun Zhou, Tianyi Li, and Yang You.
\newblock {Open-Sora:} democratizing efficient video production for all.
\newblock \emph{arXiv preprint arXiv:2412.20404}, 2024.

\bibitem[Zhu et~al.(2025)Zhu, Wang, Chen, Liu, Ye, Gu, Tian, Duan, Su, Shao, et~al.]{zhu2025internvl3}
Jinguo Zhu, Weiyun Wang, Zhe Chen, Zhaoyang Liu, Shenglong Ye, Lixin Gu, Hao Tian, Yuchen Duan, Weijie Su, Jie Shao, et~al.
\newblock {InternVL3}: Exploring advanced training and test-time recipes for open-source multimodal models.
\newblock \emph{arXiv preprint arXiv:2504.10479}, 2025.

\bibitem[Zitkovich et~al.(2023)Zitkovich, Yu, Xu, Xu, Xiao, Xia, Wu, Wohlhart, Welker, Wahid, Vuong, Vanhoucke, Tran, Soricut, Singh, Singh, Sermanet, Sanketi, Salazar, Ryoo, Reymann, Rao, Pertsch, Mordatch, Michalewski, Lu, Levine, Lee, Lee, Leal, Kuang, Kalashnikov, Julian, Joshi, Irpan, Ichter, Hsu, Herzog, Hausman, Gopalakrishnan, Fu, Florence, Finn, Dubey, Driess, Ding, Choromanski, Chen, Chebotar, Carbajal, Brown, Brohan, Arenas, and Han]{Zitkovich:RT-2}
Brianna Zitkovich, Tianhe Yu, Sichun Xu, Peng Xu, Ted Xiao, Fei Xia, Jialin Wu, Paul Wohlhart, Stefan Welker, Ayzaan Wahid, Quan Vuong, Vincent Vanhoucke, Huong~T. Tran, Radu Soricut, Anikait Singh, Jaspiar Singh, Pierre Sermanet, Pannag~R. Sanketi, Grecia Salazar, Michael~S. Ryoo, Krista Reymann, Kanishka Rao, Karl Pertsch, Igor Mordatch, Henryk Michalewski, Yao Lu, Sergey Levine, Lisa Lee, Tsang{-}Wei~Edward Lee, Isabel Leal, Yuheng Kuang, Dmitry Kalashnikov, Ryan Julian, Nikhil~J. Joshi, Alex Irpan, Brian Ichter, Jasmine Hsu, Alexander Herzog, Karol Hausman, Keerthana Gopalakrishnan, Chuyuan Fu, Pete Florence, Chelsea Finn, Kumar~Avinava Dubey, Danny Driess, Tianli Ding, Krzysztof~Marcin Choromanski, Xi~Chen, Yevgen Chebotar, Justice Carbajal, Noah Brown, Anthony Brohan, Montserrat~Gonzalez Arenas, and Kehang Han.
\newblock {RT-2:} vision-language-action models transfer web knowledge to robotic control.
\newblock In \emph{Conference on Robot Learning}, pages 2165--2183, 2023.

\end{thebibliography}

\end{document}